\documentclass[lettersize,journal]{IEEEtran}
\usepackage{amsmath,amsfonts}
\usepackage{algorithmic}
\usepackage{algorithm}
\usepackage{array}
\usepackage{textcomp}
\usepackage{stfloats}
\usepackage{url}
\usepackage{verbatim}
\usepackage{graphicx}
\usepackage{cite}
\usepackage{bibentry}
\usepackage{bm}
\usepackage{caption}
\usepackage{subcaption}
\usepackage{multirow,multicol}
\usepackage{hyperref}
\hypersetup{hidelinks,
	colorlinks=true,
	allcolors=black,
	urlcolor=red,
	pdfstartview=Fit,
	breaklinks=true}
\usepackage{amssymb}
\newcounter{MYtempeqncnt}

\begin{document}

\title{Sliding Sequential CVAE with Time Variant Socially-aware Rethinking for Trajectory Prediction}

\author{Hao Zhou, Dongchun Ren, Xu Yang, Mingyu Fan, Hai Huang
\thanks{Hao Zhou is with the National Key Laboratory of Science and Technology of Underwater Vehicle, Harbin Engineering University, Harbin, China, and also with the State Key Laboratory of Management and Control for Complex System, Institute of Automation, Chinese Academy of Sciences, Beijing, China.}
\thanks{Dongchun Ren is with the  Research Center for Autonomous Vehicles, Meituan, Beijing, China.}
\thanks{Xu Yang is with the State Key Laboratory of Management and Control for Complex System, Institute of Automation, Chinese Academy of Sciences, Beijing, China.}
\thanks{Mingyu Fan is with the College of Computer Science, Wenzhou University, Wenzhou, China, and also with the Research Center for Autonomous Vehicles, Meituan, Beijing, China.}
\thanks{Hai Huang is with the with the National Key Laboratory of Science and Technology of Underwater Vehicle, Harbin Engineering University, Harbin, China.}}

\markboth{Journal of \LaTeX\ Class Files,~Vol.~14, No.~8, August~2021}%
{Shell \MakeLowercase{\textit{et al.}}: A Sample Article Using IEEEtran.cls for IEEE Journals}

\IEEEpubid{0000--0000/00\$00.00~\copyright~2021 IEEE}

\maketitle

\begin{abstract}
Pedestrian trajectory prediction is a key technology in many applications such as video surveillance, social robot navigation, and autonomous driving, and significant progress has been made in this research topic. However, there remain two limitations of previous studies. First, with the continuation of time, the prediction error at each time step increases significantly, causing the final displacement error to be impossible to ignore. Second, the prediction results of multiple pedestrians might be impractical in the prediction horizon, \textit{i.e.}, the predicted trajectories might collide with each other. To overcome these limitations, this work proposes a novel trajectory prediction method called CSR, which consists of a cascaded conditional variational autoencoder (CVAE) module and a socially-aware regression module. The cascaded CVAE module first estimates the future trajectories in a sequential pattern. Specifically, each CVAE concatenates the past trajectories and the predicted points so far as the input and predicts the location at the following time step. Then, the socially-aware regression module generates offsets from the estimated future trajectories to produce the socially compliant final predictions, which are more reasonable and accurate results than the estimated trajectories. Moreover, considering the large model parameters of the cascaded CVAE module, a slide CVAE module is further exploited to improve the model efficiency using one shared CVAE, in a slidable manner. Experiments results demonstrate that the proposed method exhibits improvements over state-of-the-art method on the Stanford Drone Dataset (SDD) and ETH/UCY of approximately 38.0\% and 22.2\%, respectively. 
\end{abstract}


\section{Introduction}
\begin{figure}[t]
	\centering
	\includegraphics[width=\columnwidth, trim=30 10 30 30, clip]{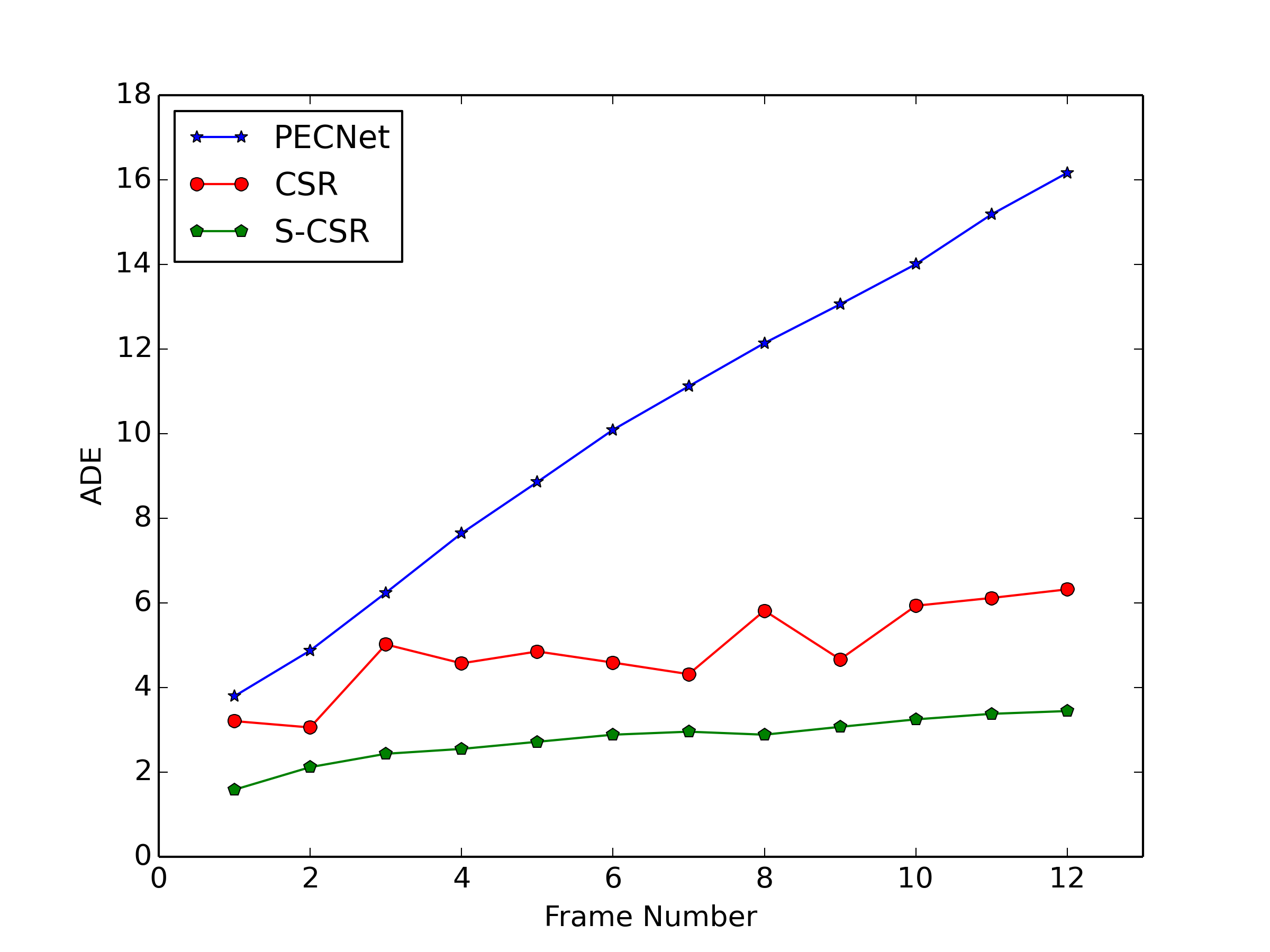} 
	\caption{The variations of the average displacement error of each frame for the PECNet, the proposed CSR and the proposed S-CSR on the SDD dataset.}
	\label{fig.1}
\end{figure}

\IEEEPARstart{T}{he} prediction of future pedestrian trajectories according to their past trajectories is a key problem for many applications such as autonomous driving \cite{HongCVPR2019, PlanningRisk2016, c18}, robotic systems \cite{vivacequa2017self, luo2018porca}, and surveillance system \cite{LVattTP2019, luber2010people}. For example, in autonomous driving scenario, a accurate prediction of pedestrians trajectories is required before the vehicle can plan a safe and effective trajectory; otherwise, a car accident may occur. 

A key challenge for pedestrian trajectory prediction is the high degree of uncertainty caused by two factors, \textit{i.e.}, the inherently multimodal attribute and the complex social interactions of pedestrians. 
In Figure \ref{fig.1}, the uncertainties of one state-of-the-art method PECNet \cite{mangalam2020not} are visualized using the average displacement error (ADE) between the ground truth and predictions. The visualization reveals that the degree of uncertainty increases over the prediction horizon. This is expected, as the movement patterns and social interactions change over time; thus, it is difficult to predict the long-term trajectories using the past trajectories. To cope with the uncertainty problem, numerous methods \cite{mangalam2020not, zhao2020tnt, mangalam2020goals} have been proposed in recent years. The core notion of these methods is to first model the target prediction, and then to predict the future trajectory conditioned on the predicted target. Despite their competitive performance, the prediction of the target point, which has a tremendous influence on the final prediction, is inherently a  difficult problem that has the maximum degree of uncertainty. For example, two different target points will lead to the generation of two completely different trajectories.

In Figure \ref{fig.1}, the uncertainty can also be analyzed from another perspective, namely that the first prediction time step has the minimum degree of uncertainty; thus, it can be easily captured by the past trajectory and has the best prediction accuracy. This is also expected because the first prediction time step has a movement pattern similar to that of the past trajectories. Accordingly, in this paper, interest has been placed in the development of a model that can sequentially predict the following time step using the updated past trajectory as the input. Recurrent Neural Networks (RNN)-based sequential methods \cite{alahi2016social, gupta2018social, sadeghian2019sophie, liang2019peeking, xu2020cf} are similar to the idea explored in this study. They use the hidden states of the predicted points before the prediction of the next point as the input. However, the hidden state feature of the decoder cannot explicitly encode the predicted points as the past trajectories.

\begin{figure*}[htp]
	\centering
	\includegraphics[width=0.98\textwidth, trim=20 10 20 10, clip]{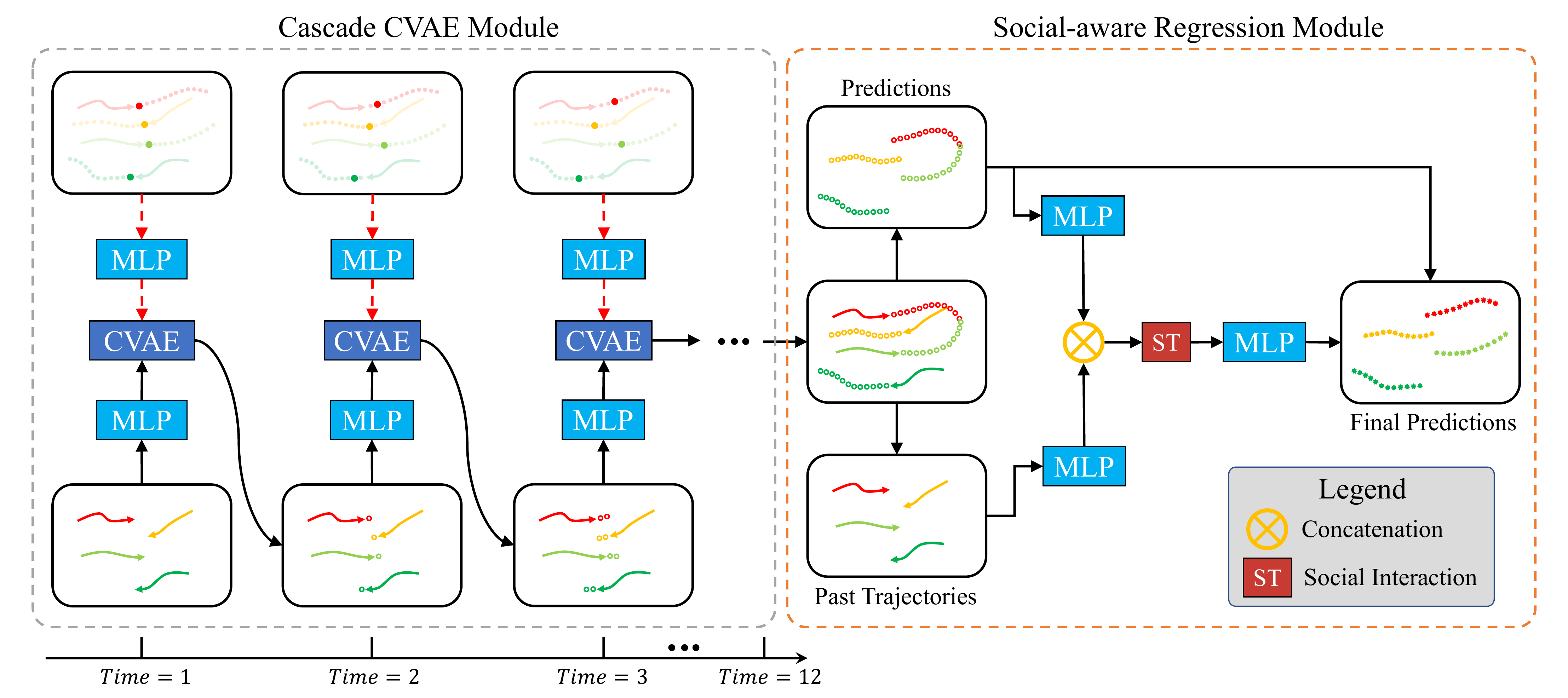} 
	\caption{The overview of the proposed CSR model. First, the cascaded CVAE module estimates the future trajectory in a sequential manner. Specifically, each CVAE concatenates the past trajectory and the predicted points so far as the input and predicts the location at the following time step. Then, the socially-aware regression module extracts interaction features to refine the predicted trajectory.}
	\label{fig.2}
\end{figure*}

This work presents a novel cascaded conditional variational autoencoder (CVAE) with socially-aware rethinking method, as illustrated in Figure \ref{fig.2}. The proposed method consists of two modules, namely the cascaded CVAE module and the socially-aware regression module. In the cascaded CVAE module, unshared CVAEs are used to sequentially predict the future points of different time steps using the updated past trajectories as the input. The past trajectory after each prediction is updated by concatenating it with the newly predicted point at the current time. Via this strategy, the predicted points are treated as pseudo-labels to decrease the degree of prediction uncertainty of the next point, thereby improving the long-term trajectory prediction performance. The average displacement error of the proposed method is also visualized in Figure \ref{fig.1}, from which it is evident that the proposed method is characterized by the decreases of the average displacement errors of long-term future points. 

Moreover, the improvement is also partially attributed to the socially-aware regression module. The future trajectories predicted by the cascaded CVAE module may be socially unacceptable because the CVAE can not model interactions among pedestrians. Thus, the socially-aware regression module is proposed to refine the predicted trajectories via social interactions. This module first extracts features of the past trajectories and the predicted trajectories by two encoders. Then the two features are concatenated and fed to a social attention mechanism to extract the interaction feature. Finally, the decoder decodes the interaction feature to produce the offsets of the future trajectories, which are then used to refine the future trajectories predicted by the cascaded CVAE module.

In the cascaded CVAE module, $\delta$ unshared CVAEs are used to predicted $\delta$ future points using incremental inputs. Thus, the proposed module has large model parameters and slow inference speed. Actually, the predictions of different future points fit the independent and identically distributed. Thus, one CVAE is enough to predict $\delta$ future points. Based on this observation, a more lightweighted slide CVAE module is introduced to adopt one shared CVAE to predict future points using the updated past trajectory of fixed length, in a slidable manner, as illustrated in Figure \ref{Fig.5c}. Moreover, the proposed slide CVAE module is integrated with the socially-aware regression module resulting in an more efficiency end-to-end trajectory predicition method, called the S-CSR. The average displacement error of the S-CSR is also visualized in Figure \ref{fig.1}, from which it is indicates that the slide CVAE module can futher improve the prediction accuracy of the long-term future points.

The contributions are summarized as follows:

\begin{itemize}
	\item A cascaded CVAE with socially-aware rethinking method is proposed for pedestrian trajectory prediction. The proposed method consists of the cascaded CVAE module or slide CVAE module and the socially-aware regression module. 
	\item The proposed cascaded CVAE module uses unshared CVAEs to sequentially predict the future points using the updated past trajectories as the input.
	\item The proposed socially-aware regression module extracts interaction features to generate offsets to refine the future trajectories predicted by the cascaded CVAE module.
	\item The proposed slide CVAE module uses one shared CVAE to predict the future points using the updated past trajectory of fixed length, in a slidable manner.
	\item  The results of extensive experiments demonstrate that the proposed CSR and S-CSR surpass state-of-the-art methods by a large margin on two benchmark datasets.
\end{itemize}

\section{Related Work}

\subsection{Direct Prediction Methods}
The direct strategy involves the utilization of one model that is capable of predicting the entire future trajectory in a one-shot manner. Most trajectory prediction methods are characterized by direct prediction. For example, early works adopted the Bayesian network method \cite{lefevre2011exploiting}, the Gaussian process regression method \cite{ramussen2006gaussian}, and the kinematic method \cite{toledo2009imm} to directly predict the future sequence. However, these method did not consider the interactions between agents, and usually failed in crowded scenarios.

Recently, with the development of deep learning, convolutional neural network (CNNs) and multilayer perceptron (MLPs) have been used in trajectory prediction. For example, GRIP \cite{li2019grip} uses a temporal graph to model the interactions and a temporal CNN to predict the future sequences. VectorNet \cite{gao2020vectornet} utilizes self-attention to aggregate interactions and the MLP to predict future sequences. Social-STGCN \cite{mohamed2020social} and AST-GNN \cite{zhou2021ast} capture interactions using a graph attention (GAT) and predict future sequences using a temporal CNN. 

Albert it is simple and efficient, the direct strategy uses only the past trajectory for prediction, but has no opportunity to extract useful information from the predicted future points. In contrast, the proposed method uses the predicted points to update the past trajectory, thereby improving the prediction accuracy of the following time step.

\subsection{Recursive Prediction Methods}
The recursive strategy involves the use of a one-step model multiple times, in which the predictions for the prior time steps are used for the prediction of the following time step. RNN models including long short-term memory (LSTM) \cite{hochreiter1997long} and gated recurrent unit (GRU) \cite{chung2014empirical} have been widely adopted in the recursive prediction methods. For example, S-LSTM \cite{alahi2016social} uses LSTM to extract the past trajectories, and then aggregates the interactions of different pedestrians in the hidden state using a social pooling mechanism. CF-LSTM \cite{xu2020cf} improves LSTM by employing a cascaded-feature that can simultaneously capture location and velocity information. S-GAN \cite{gupta2018social} extends S-LSTM by using GAN to generate multimodal predictions. SoPhie \cite{sadeghian2019sophie} and PIF \cite{liang2019peeking} use a CNN to extract scene features and LSTM to extract motion features, and then adopt LSTM to predict scene-compliant trajectories. Trajectron++ \cite{salzmann2020trajectron++} introduces a modular graph-structured recurrent model that can integrate agent dynamics and heterogeneous data into trajectory prediction. Although the hidden state of the predicted poins has been adopted to assist in the prediction of the following time step, it does not explicitly model the predicted points as the past trajectories.

\subsection{Target-conditioned Methods}
The target-conditioned strategy first predicts the target points and then generates the future trajectories conditioned on the targets. For example, PECNet \cite{mangalam2020not} first predicts distant trajectory endpoints using a CVAE and then generates multimodal predictions conditioned on the endpoints. TNT \cite{zhao2020tnt} improves PECNet by introducing a scoring and selection stage that ranks and selects multimodal predictions with likelihood scores. Y-net \cite{mangalam2020goals} further improves PECNet by iteratively predicting intermediate waypoints and trajectories. The performance of target-conditioned methods is easily affected by the predicted targets, however, target prediction is inherently a difficult problem with a high degree of uncertainty.

\section{Methodology}
Pedestrian trajectory prediction is a task to predict the future trajectories of pedestrians based on their past trajectories. First, it is assumed that videos of pedestrians walking are preprocessed to obtain the spatial coordinates of pedestrians. For example, $p_i^t = \left(x_i^t, y_i^t\right)$ represents the coordinate of the $i$-th pedestrian at the $t$-th time step. Then, the coordinates of each pedestrian in the scene are classified into the past and future trajectories according to two default values, namely the observation horizon $\tau$ and the prediction horizon $\delta$. The definitions of the past trajectory $\mathbf{X}_i$ and the future trajectory $\mathbf{Y}_i$ are as follows:
\begin{equation}
	\mathbf{X}_i = \left\{\left(x_i^t, y_i^t\right) | \forall t \in \left\{1, 2, ..., \tau \right\}\right\},
\end{equation}
\begin{equation}
	\mathbf{Y}_i = \left\{\left(x_i^t, y_i^t\right) | \forall t \in \left\{\tau+1, \tau+2, ..., \tau+\delta \right\}\right\}.
\end{equation}
Finally, given the past trajectories $\left\{\mathbf{X}_i |\forall i \in \left\{1, 2, ..., N\right\} \right\}$ of pedestrians in scene, the goal is to generate future predictions  $\left\{\hat{\mathbf{Y}}_i |\forall i \in \left\{1, 2, ..., N\right\} \right\}$ that minimize the average displacement error with the ground truth future trajectories  $\left\{\mathbf{Y}_i |\forall i \in \left\{1, 2, ..., N\right\} \right\}$, where $N$ represents the number of pedestrians in a scene. The definition of $\hat{\mathbf{Y}}_i$ is as follows:
\begin{equation}
	\hat{\mathbf{Y}}_i = \left\{\left(\hat{x}_i^t, \hat{y}_i^t\right)| \forall t \in \left\{\tau+1, \tau+2, ..., \tau+\delta \right\} \right\}.
\end{equation}

To generate reliable predictions, a pedestrian trajectory prediction method that consists of a cascaded CVAE module and a socially-aware regression module is proposed, as shown in Figure \ref{fig.2}. Because the direct prediction of long-range future trajectory $\mathbf{Y}_i$ has a high degree of uncertainty, a cascaded CVAE module is first designed to predict the future points in a sequential manner using the updated past trajectoies as the input. Then, considering that the CVAE is unable to model social interactions, a socially-aware regression module is proposed to further refine the predicted future trajectories using social interaction knowledge.

\begin{figure}[t]
	\centering
	\includegraphics[width=0.98\columnwidth,  trim=10 10 10 10, clip]{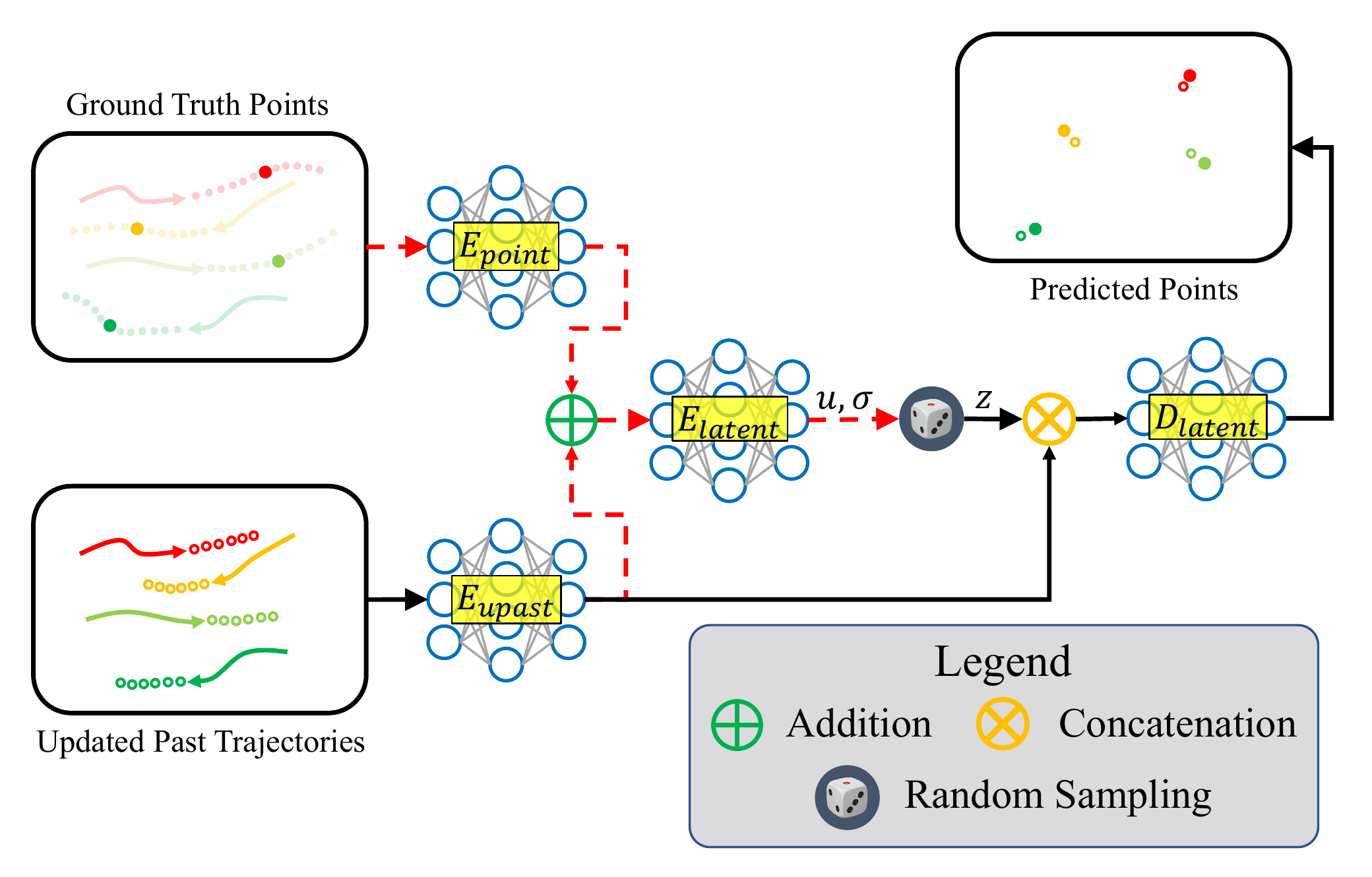} 
	\caption{The illustration of the cascaded CVAE unit. The CVAE uses the updated past trajectory and ground truth future point to train a model for multimodal future point prediction. The dots and circles respectively represent the predicted future points and the ground truth future points. Red dotted lines denote the layers utilized only during training.}
	\label{fig.3}
\end{figure}

\subsection{Cascaded CVAE Module}

The prediction of the future trajectories is inherently ambiguous and has a high degree of uncertainty. Thus, it is difficult to learn a deterministic function $f$ that directly maps the past trajectories $\left\{\mathbf{X}_i | \forall i \in \left\{ 1, 2, ..., N \right\}\right\}$ to the future trajectories $\left\{\mathbf{Y}_i | \forall i \in \left\{ 1, 2, ..., N \right\} \right\}$. 

However, as described in the Introdction, the degrees of the uncertainty of the future points are increase with the continuation of time, and the first future point has the minimum degree of uncertainty. According to this phenomenon, the use of cascaded CVAEs is proposed to decompose the prediction of the future trajectories into the prediction of the future points using the updated past trajectories, in a sequential manner, as illustrated in Figures \ref{fig.2} and \ref{Fig.5b}. 

The detailed structure of each cascaded CVAE is illustrated in Figure \ref{fig.3}. The CVAE infers a distribution of future points based on the updated past trajectory. The training of the CVAE requires two inputs, namely the updated past trajectory $\widetilde{\mathbf{X}}_i^t$ and the ground point $p_i^t=\left(x_i^t, y_i^t\right)$ of pedestrian $i$ at future time step $t$.  The updated past trajectory consists of the past trajectory and the predicted future points before time step $t$ using the concatenation operation, such that:
\begin{equation}
	\widetilde{\mathbf{X}}_i^t = Concat\left(\mathbf{x}_i , \left\{\left(\hat{x}_i^s, \hat{y}_i^s\right) | \forall s \in \left\{ \tau+1, \tau+2, ..., t-1 \right\}\right\}\right).
\end{equation} 
The updated past trajectory $\widetilde{\mathbf{X}}_i^t$ and the ground truth future point $p_i^t$ are first encoded using two MLP encoders $E_{upast}$ and $E_{point}$, as follows:
\begin{equation}
	\mathbf{f}_{upast} = E_{upast}\left(\widetilde{\mathbf{X}}_i^t\right),
\end{equation}
\begin{equation}
	\mathbf{f}_{point} = E_{point}\left(p_i^t\right).
\end{equation}
Then, the two feature representations $\mathbf{f}_{upast}$ and $\mathbf{f}_{point}$ are concatenated together and fed to the MLP encoder $E_{latent}$ to generate parameters $\left(\bm{\mu}_i^t, \bm{\sigma}_i^t\right)$ of the latent distribution, such that:
\begin{equation}
	\left(\bm{\mu}_i^t, \bm{\sigma}_i^t\right) = E_{latent}\left(Concat\left(\mathbf{f}_{upast}, \mathbf{f}_{point}\right)\right).
\end{equation}
Finally, the latent variable $\mathbf{z}_i^t$ is randomly sampled from distribution $\mathcal{N}\left(\bm{\mu}_i^t, \bm{\sigma}_i^t\right)$, and is concatenated with the feature $\mathbf{f}_{upast}$ and yield prediction $\hat{p}_i^t=\left(\hat{x}_i^t, \hat{y}_i^t\right)$ using the MLP decoder $D_{latent}$, as follows:
\begin{equation}
	\hat{p}_i^t = D_{latent}\left(Concat\left(\mathbf{z}_i^t, \mathbf{f}_{point}\right)\right).
\end{equation}
In the inference stage, because the ground truth point $p_i^t$ is unavailable, the latent variable $\mathbf{z}_i^t$ is randomly sampled from a prior distribution $\mathcal{N}\left(0, \mathbf{I}\right)$, and concatenated with the feature $\mathbf{f}_{upast}$, and the prediction $\hat{p}_i^t$ is then generated using the trained decoder $D_{latent}$.

For each pedestrian $i$, there are $\delta$ future time steps. Thus, $\delta$ CVAEs with the same structure are sequentially operated to generate the predictions of $\delta$ future points $\hat{p}_i^{\tau+1}, \hat{p}_i^{\tau+2}, ..., \hat{p}_i^{\tau+\delta}$. It should be noted that model parameters for the CVAE, including $E_{upast}, E_{point}, E_{latent}$ and $D_{latent}$, are not shared because the length of the updated past trajectory of each CVAE is different, and the CVAEs are used to predict future points at different future time steps.

\begin{figure}[!t]
	\centering
	\includegraphics[width=0.98\columnwidth,  trim=15 10 15 15, clip]{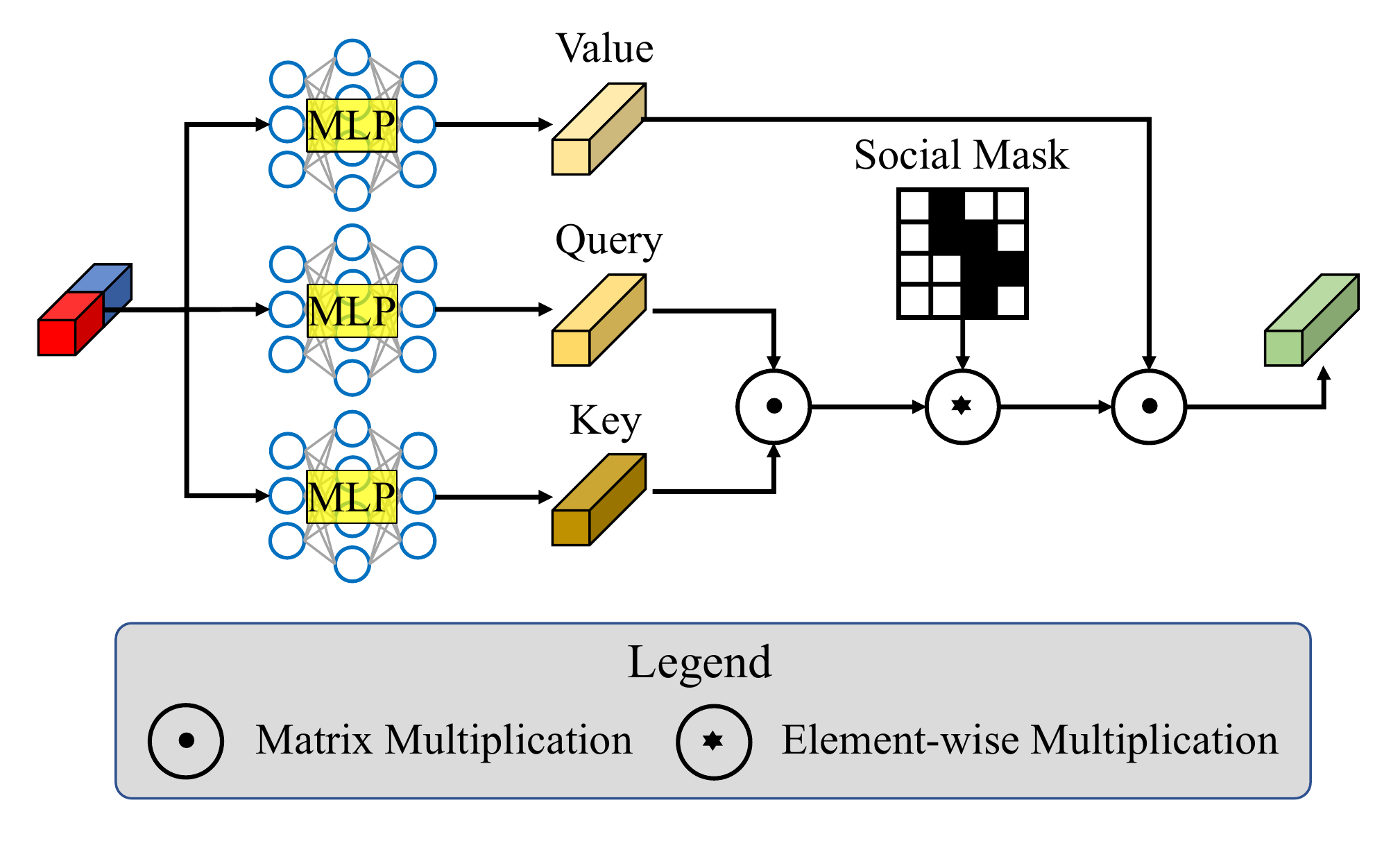} 
	\caption{The illustration of the social non-local pooling layer proposed by \cite{mangalam2020not}.}
	\label{fig.4}
\end{figure}

\subsection{Socially-aware Regression Module}

The CVAE cannot model interactions among different pedestrians; thus, the future trajectories predicted by the cascaded CVAE module may rough and socially unacceptable. To cope with this problem, a socially-aware regression module, as illustrated in Figure \ref{fig.2}, is proposed to estimate the offsets to refine the predicted future trajectories.

To generate effective offsets, two MLP encoders $E_{opast}$ and $E_{pfuture}$ are first used to extract features from the original past trajectory and the predicted future trajectory, as follows:
\begin{equation}
	\mathbf{f}_{opast} = E_{opast}\left(\mathbf{X}_i\right),
\end{equation}
\begin{equation}
	\mathbf{f}_{pfuture} = E_{pfuture}\left(\hat{\mathbf{Y}}_i\right).
\end{equation}
Then, the two feature representations $\mathbf{f}_{opast}$ and $\mathbf{f}_{pfuture}$ are concatenated and passed into a social interaction layer $SE$ to produce an interaction feature, such that:
\begin{equation}
	\mathbf{f}_{st} = SE\left(Concat\left(\mathbf{f}_{opast}, \mathbf{f}_{pfuture}\right)\right).
\end{equation}
Finally, the extracted feature $\mathbf{f}_{st}$ is fed to the MLP decoder $D_{offsets}$ to estimate the offsets of the predicted future trajectories:
\begin{equation}
	\Delta\hat{\mathbf{Y}}_i = D_{offsets}\left(\mathbf{f}_{st}\right),
\end{equation}
where $\Delta\hat{\mathbf{Y}}_i = \left\{\left(\Delta\hat{x}_i^t, \Delta\hat{y}_i^t\right) | \forall t \in \left\{ \tau+1, \tau+2, ..., \tau+\delta \right\} \right\}$. Using the estimated offsets, the predictions $\hat{\mathbf{Y}}_i$ is refined by $\hat{\mathbf{Y}}_i + \Delta{\hat{\mathbf{Y}}_i}$.

In this paper, the social non-local pooling layer proposed by \cite{mangalam2020not} is adopted as the $SE$ layer. This method integrates the social pooling \cite{alahi2016social} into the non-local attention mechanism \cite{wang2018non}, as illustrated in Figure \ref{fig.4}. Specifically, three MLPs are first used to extract the query, key, and value features. Then, a matrix multiplication is applied between the query and key features to produce an attention map. Note that the social mask is a binary matrix calculated using the distance between the pedestrians. Next, a social mask is used point-wise on the attention map to mask out the non-consecutive neighbors in the scene. Finally, a weighted sum between the masked attention map and the value feature is conducted to generate a social interaction feature.

\begin{figure}[!t]
	\centering
	\subfloat[Baseline method]{
		\begin{minipage}[b]{0.49\textwidth}
			\includegraphics[width=1\textwidth, , trim=0 8 0 5, clip]{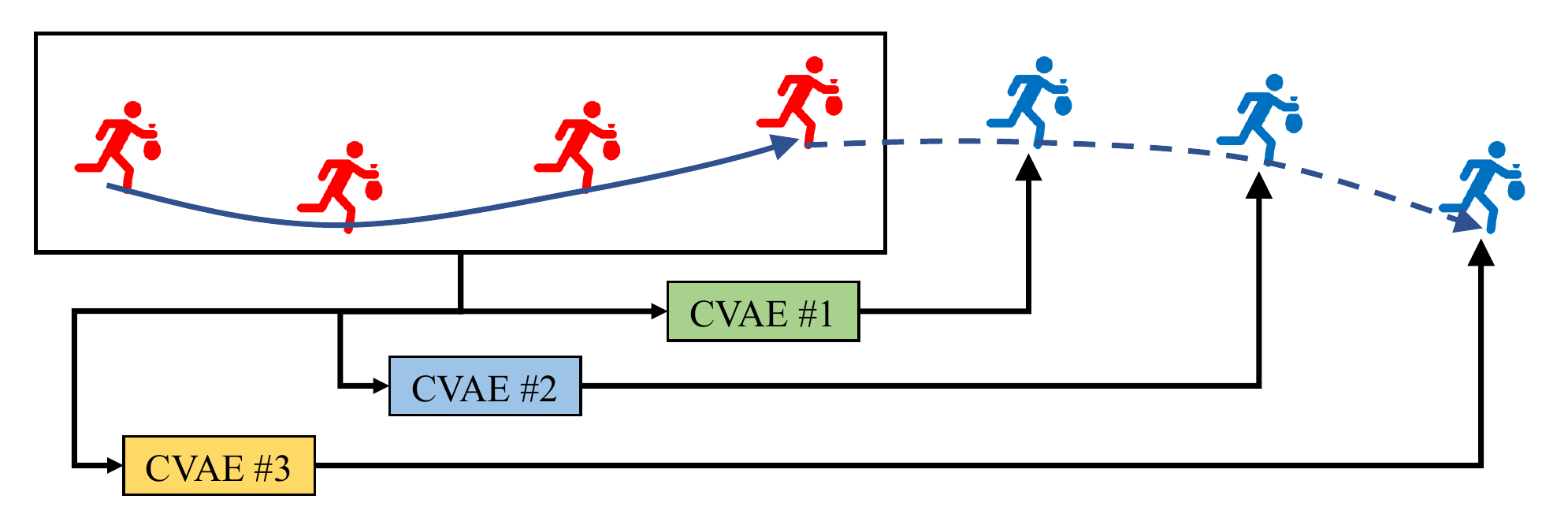} 
		\end{minipage}
		\label{Fig.5a}
	}
	\\ 
	\subfloat[Cascaded CVAE module]{
		\begin{minipage}[b]{0.49\textwidth}
			\includegraphics[width=1\textwidth, trim=0 8 0 5, clip]{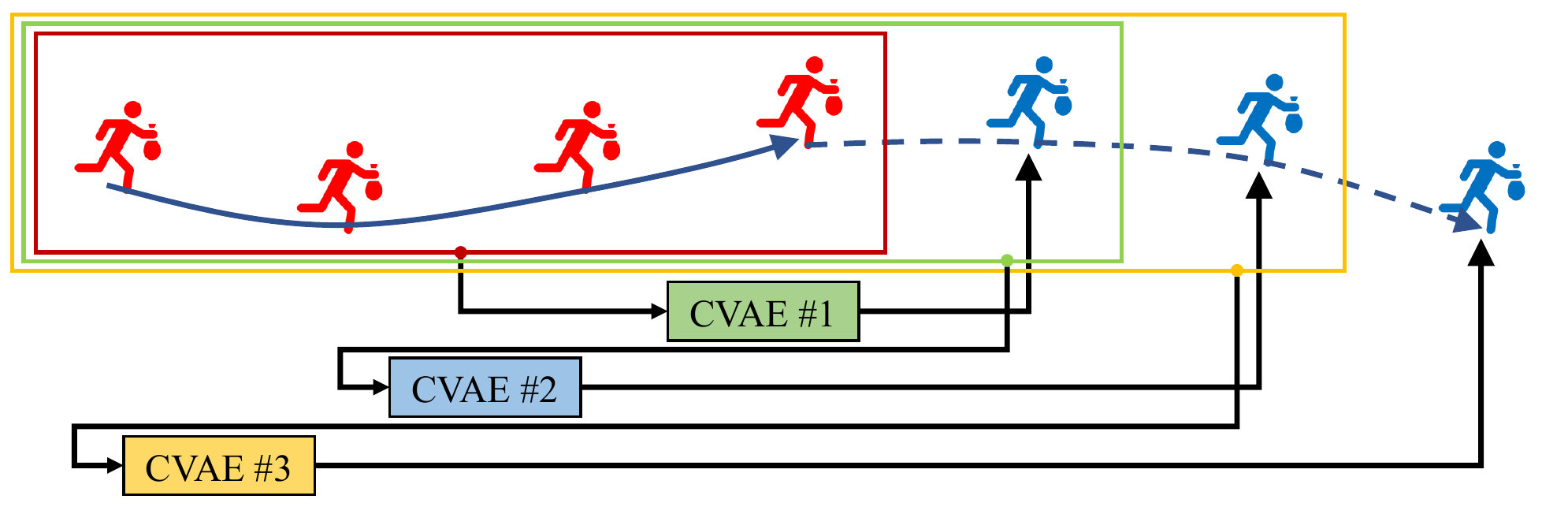} 
		\end{minipage}
		\label{Fig.5b}
	}
	\\
	\subfloat[Slide CVAE module]{
		\begin{minipage}[b]{0.49\textwidth}
			\includegraphics[width=1\textwidth, , trim=0 8 0 5, clip]{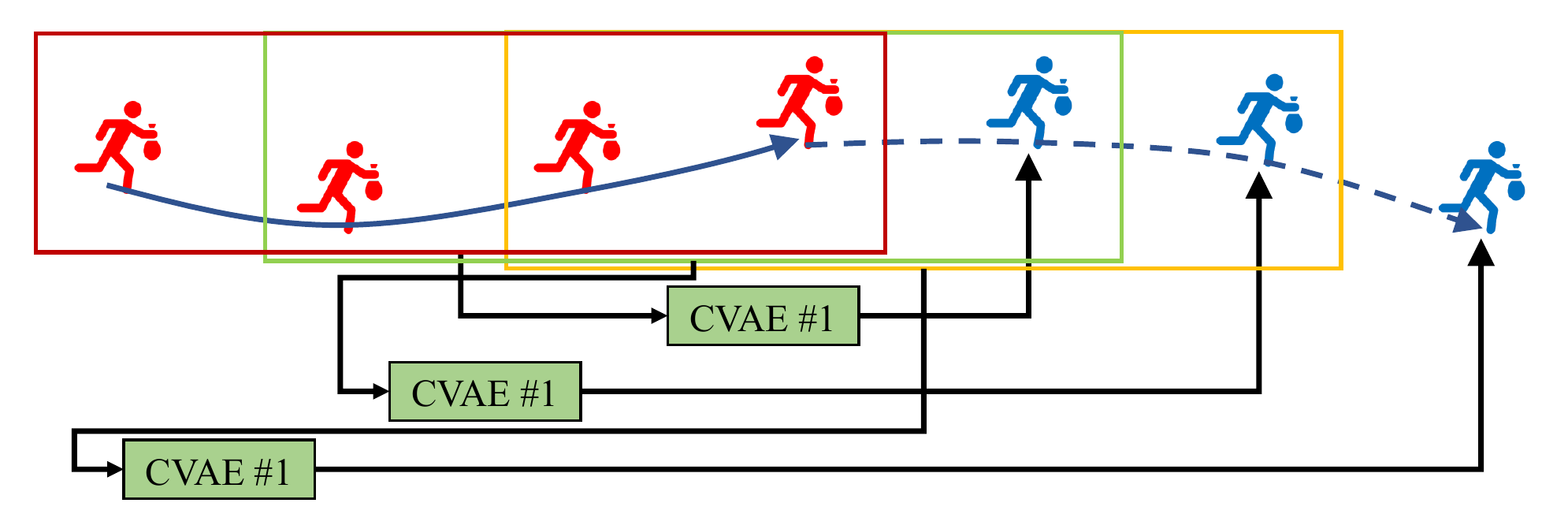} 
		\end{minipage}
		\label{Fig.5c}
	}
	\caption{The illustration of the baseline method, the cascaded CVAE module and the slide CVAE module.}
	\label{Fig.5}
\end{figure}

\begin{figure*}[!t]
	\normalsize
	\setcounter{MYtempeqncnt}{\value{equation}}
	\setcounter{equation}{12}
	\begin{equation}
		\overline{\mathbf{X}}_i^t = \left\{
		\begin{array}{cc}
			Concat\left(\left\{\left(x_i^s, y_i^s\right) | \forall t \in \left\{t-\alpha,  ..., \tau \right\}\right\}, \left\{\left(\hat{x}_i^s, \hat{y}_i^s\right) | \forall s \in \left\{ \tau+1, \tau+2, ..., t-1 \right\}\right\}\right) &  {\tau + 1 \leq t \leq \tau + \alpha}\\
			Concat\left(\left\{\left(\hat{x}_i^s, \hat{y}_i^s\right) | \forall s \in \left\{ \tau+1, \tau+2, ..., t-1 \right\}\right\}\right) &  {\tau + \alpha + 1 \leq t \leq \tau+\delta}\\
		\end{array}
		\right. 
		\label{eq.13}
	\end{equation}
	\setcounter{MYtempeqncnt}{\value{equation}}
	\hrulefill
\end{figure*}

\subsection{Slide CVAE Module}
Even though the updated past trajectory that contains the predicted future points before current time step is adopted by cascade CVAE module to assist in the predicting of the future point at the following time step. There exists two drawbacks of the proposed module. First, $\delta$ CVAEs are need to predict $\delta$ future points, resulting in large model parameters. Second, with the continuation of time, the updated past trajectory is increase, and thus decrease the inference speed of CVAE. 
		
Actually, the predictions of different future points fit the independent and identically distributed. Thus, one CVAE is enough to predict $\delta$ future points. Based on this observation, different from the cascaded CVAE module, a slide CVAE module is proposed to decompose the prediction of the future trajectories into the prediction of the future points using the updated past trajectories of fixed length, in a slidable manner, as illustrated in Figure \ref{Fig.5c}. In this manner, the model parameters are decreased by using only one CVAE, and the inference speed is increased by using the updated past trajectory of fixed length.
		
The proposed slide CVAE has the same structure of each cascaded CVAE, except for the input. The slide CVAE module uses the updated trajectories of fixed length. Specifically, at the time step $t$, the input of slide CVAE is consists of the original past trajectory and the predicted points before current time step.  To keep the fixed length of the updated past trajectory, the input $\overline{\mathbf{X}}_i^t$ of the slide CVAE module is defined as Equation \ref{eq.13}, where $\alpha$ is the fixed length of the updated past trajectory.

\subsection{Training the Model}

To train the $\delta$ CVAEs in the cascaded CVAE module, two loss terms, namely the average point loss $\mathcal{L}_{AP}$  and the  Kullback-Leibler (KL) divergence loss $\mathcal{L}_{KLD}$, are used. The definitions of $\mathcal{L}_{AP}$ and $\mathcal{L}_{KLD}$ are as follows:
\begin{equation}
	\mathcal{L}_{AP} = \frac{1}{N}\sum_{i=1}^{N}\sum_{t=\tau+1}^{\tau+\delta} \left \|p_i^t - \hat{p}_i^t \right \|^2,
\end{equation}
\begin{equation}
	\mathcal{L}_{KLD} = \frac{1}{N}\sum_{i=1}^{N}\sum_{t=\tau+1}^{\tau+\delta}D_{KL}\left(\mathcal{N}\left(\bm{\mu}_i^t, \bm{\sigma}_i^t\right) || \mathcal{N}\left(0, \mathbf{I}\right) \right).
\end{equation}
The KL divergence loss is a regularization loss that measures the distance between the sampling distribution at the test stage and the sampling distribution of the latent variable learned at the training stage, and is used to train the CVAEs. The average point loss measures the displacement error between the predicted future points and the ground truth future points, and is used to train $E_{upath}$, $E_{point}$, $E_{latent}$, and $D_{latent}$. In addition, a regression loss $\mathcal{L}_R$ is used to train the socially-aware regression module. The definition of $\mathcal{L}_R$ is:
\begin{equation}
	\mathcal{L}_{R} = \frac{1}{N}\sum_{i=1}^{N}\sum_{t=\tau+1}^{\tau+\delta}\left\| p_i^t - \hat{p}_i^t -\Delta \hat{p}_i^t\right\|,
\end{equation}
where $\Delta p_i^t=\left(\Delta \hat{x}_i^t, \Delta \hat{y}_i^t\right)$. The regression loss measures the displacement between the predicted offsets and the real offsets, and is used to train $E_{opast}$, $E_{pfuture}$, and $D_{offsets}$.  Finally, the total loss $\mathcal{L}_T$ is defined as follows:
\begin{equation}
	\mathcal{L}_{T} = \mathcal{L}_{KLD} + \mathcal{L}_{AP} + \mathcal{L}_{R}. 
\end{equation}
This multi-task loss is used to train the entire model in an end-to-end manner.

\section{Experiments}
\subsection{Datasets}
Two pedestrian trajectory prediction datasets, namely the Stanford Drone Dataset (SDD) and the ETH/UCY dataset, were used to validate the proposed method. The trajectories in both datasets were sampled with a frame rate of 2.5Hz; the former 3.2 seconds of trajectories (eight frames) are used as the inputs, and the latter 4.8 seconds (twelve frames) are used as the future trajectories to be predicted.

\textbf{The SDD} \cite{robicquet2016learning} is an aerial-view pedestrian trajectory prediction dataset captured by a drone. The dataset contains eight scenes of a college campus. In total, there are totally 11000 pedestrians with 185000 interactions between agents and 40000 interactions between agents and scenes \cite{robicquet2016learning}. The standard method for the division of the training and test sets used in \cite{sadeghian2019sophie, gupta2018social, deo2020trajectory} was adopted in the experiments. 

\textbf{The ETH/UCY} \cite{pellegrini2009you, lerner2007crowds} is a dataset group that contains two datasets, \textit{i.e.} ETH and UCY datasets. The ETH and UCY datasets contains five scenes, namely ETH, HOTEL, ZARA1, ZARA2, and UNIV. In these scenes, there are a total of 1536 unique pedestrians and thousands of nonlinear trajectories. 
To ensure fair comparisons, the same leave-one-out strategy adopted in previous studies \cite{alahi2016social, gupta2018social, kosaraju2019social, sadeghian2019sophie} was used to train and evaluate the proposed model. 

\subsection{Implementation Details}
The proposed CSR was implemented in PyTorch \cite{paszke2019pytorch} installed on a desktop computer ran the Ubuntu 16.04 operating system and contained an NVIDIA 1080TI GPU. The Adam algorithm \cite{kingma2014adam} with a learning rate of $3e^{-4}$ was used as the optimizer to train the CSR model. In each experiment, the model was trained for 600 epochs with a batch size 512.

\begin{table}[h]
	\begin{center}
		\begin{tabular}{|c|c|}
			\hline
			Name & Network Architecture \\
			\hline
			$E_{upast}$ & $2t \times 512 \times 256 \times 16 \left(\tau \le t \le \tau+\delta-1\right) $ \\ 
			\hline
			$E_{point}$ & $2 \times 8 \times 16 \times 16$ \\ 
			\hline
			$E_{latent}$ & $32 \times 8 \times 50 \times 32$ \\
			\hline
			$D_{latent}$ & $32 \times 1024 \times 512 \times 1024 \times 2$\\
			\hline
			$E_{opast}$ & $16 \times 512 \times 256 \times 16 $\\
			\hline
			$E_{future}$ & $24 \times 512 \times 256 \times 16 $ \\
			\hline
			$D_{offsets}$ & $ 32 \times 1024 \times 512 \times 1024 \times 24$\\
			\hline
		\end{tabular}
	\end{center}
	\caption{The detailed architectures of all sub-networks.}
	\label{tab.1}
\end{table}

\begin{table*}
	\centering
	\begin{subtable}[t]{\linewidth}
		\centering
		\begin{tabular}{c||c|c|c|c|c|c||c|c}
			\hline
			Methods & S-GAN & CF-VAE & P2TIRL & SimAug & PECNet & Y-net$^\star$ & CSR & S-CSR \\
			\hline
			ADE & 27.23 & 12.60 & 12.58 & 10.27 & 9.96 & 7.85 & 4.87 & \textbf{2. 77} \\
			\hline
			FDE & 41.44 & 22.30 & 22.07 & 19.71 & 15.88 & 11.85 & 6.32 & \textbf{3. 45}\\
			\hline
		\end{tabular}
		\label{tab.2a}
		\caption{Best of 20 samples}
	\end{subtable}
	
	\qquad
	
	\begin{subtable}[t]{\linewidth}
		\centering
		\begin{tabular}{ c||c|c|c|c||c|c}
			\hline
			Methods & DESIRE & TNT & PECNet & Y-net$^\star$ & CSR & S-CSR \\
			\hline
			ADE & 19.25 & 12.79  & 12.23 & 11.49 & 8.38 & \textbf{6.02}\\
			\hline
			FDE & 34.05 & 29.58 & 21.16 & 20.23 & 13.43 & \textbf{8.74}\\
			\hline
		\end{tabular}
		\label{tab.2b}
		\caption{Best of 5 samples}
	\end{subtable}
	\caption{The comparison of the proposed CSR and S-CSR models with several baseline methods and previous state-of-the-art methods (labeled by $^\star$) on the SDD dataset. (a) Results are tested using best of 5 samples. (b) Results are tested using best of 20 samples. The best results are highlighted in bold.}
	\label{tab.2} 
\end{table*}

\begin{table*}[h]
	\centering
	\resizebox{\linewidth}{!}{
		\begin{tabular}{c||c|c|c|c|c|c|c|c|c|c|c|c|c|c||c|c|c|c}
			\hline
			Scenes & \multicolumn{2}{|c|}{S-LSTM} & \multicolumn{2}{|c|}{S-GAN} & \multicolumn{2}{|c|}{Sophie} & \multicolumn{2}{|c|}{CGNS} & \multicolumn{2}{|c|}{PECNet} & \multicolumn{2}{|c|}{Trajectron++} & \multicolumn{2}{|c||}{Y-net$^\star$} & \multicolumn{2}{|c|}{CSR} & \multicolumn{2}{|c}{S-CSR}\\
			\hline
			Metrics & ADE & FDE & ADE & FDE & ADE & FDE & ADE & FDE & ADE & FDE & ADE & FDE & ADE & FDE & ADE & FDE & ADE & FDE \\
			\hline
			ETH & 0.70 & 1.40 & 0.81 & 1.52 & 0.70 & 1.43 & 0.62 & 1.40 & 0.54 & 0.87 & 0.39 & 0.83 & 0.28 & \textbf{0.33} & 0.28 & 0.53 & \textbf{0.19} & 0.35\\
			\hline
			HOTEL & 0.37 & 0.73 & 0.72 & 1.61 & 0.76 & 1.67& 0.70 & 0.93 & 0.18 & 0.24 & 0.12 & 0.21 & 0.10 & 0.14 & 0.07 & 0.08 & \textbf{0.06} & \textbf{0.07}\\
			\hline
			UNIV & 0.60 & 1.32 & 0.60 & 1.26 & 0.54 & 1.24 & 0.48 & 1.22 & 0.35 & 0.60 & 0.20 & 0.44 & 0.24 & 0.41 & 0.24 & 0.35 & \textbf{0.13} & \textbf{0.21}\\
			\hline
			ZARA1 & 0.49 & 1.15 & 0.34 & 0.69 & 0.30 & 0.63 & 0.32 & 0.59 & 0.22 & 0.39 & 0.15 & 0.33 & 0.17 & 0.27 & 0.07 & 0.09 & \textbf{0.06} & \textbf{0.07}\\
			\hline
			ZARA2 & 0.39 & 0.89 & 0.42 & 0.84 & 0.38 & 0.78 & 0.35 & 0.71 & 0.17 & 0.30 & 0.11 & 0.25 & 0.13 & 0.22 & \textbf{0.05} & 0.09 & \textbf{0.05} & \textbf{0.08}\\
			\hline
			AVG & 0.51 & 1.10 & 0.58 & 1.18 & 0.54 & 1.15 & 0.49 & 0.97 & 0.29 & 0.48 & 0.19 & 0.41 & 0.18 & 0.27 & 0.14 & 0.23 & \textbf{0.10} & \textbf{0.16}\\
			\hline
		\end{tabular}
	}
	\caption{The comparison of the proposed CSR and S-CSR models with several baseline methods and previous state-of-the-art methods (labeled by $^\star$) on the ETH/UCY dataset. Results are tested using best of 20 samples. The best results are highlighted in bold.}
	\label{tab.3}
\end{table*}

\textbf{Network Configuration:} The sub-networks of the two proposed modules are consists of the MLP and ReLU activation functions. The detailed network architectures of these sub-networks are presented in Table \ref{tab.1}.

\textbf{Evaluation Metrics:} To evalulate the proposed method, two commonly used evaluation metrics, namely the average displacement error (ADE) and the final displacement error (FDE) were employed. The definitions of the ADE and FDE are as follows:
\begin{equation}
	ADE = \frac{\sum_{i=1}^{N}\sum_{t=\tau+1}^{\tau+\delta}\left\|p_i^t - \hat{p}_i^t\right\|}{N\delta},
\end{equation}
\begin{equation}
	FDE = \frac{\sum_{i=1}^{N}\left \| p_i^{\tau+\delta} - \hat{p}_i^{\tau+\delta}\right \|}{N}.
\end{equation}
The ADE measures the mean Euclidean distance between the predicted future points and the ground truth future points, and the FDE measures the Euclidean distance between the predicted final points and the ground truth final points.

\textbf{Baselines:} The proposed method was compared with the following state-of-the-art baseline methods:
\begin{itemize}
	\item  S-LSTM \cite{alahi2016social} Each pedestrian is modeled using an LSTM, and the interactions are extracted by pooling the hidden states among neighbors.
	\item DESIRE \cite{lee2017desire}. An inverse optimal control (IOC)-based trajectory planning method is used to refine the predicted trajectories.
	\item S-GAN \cite{gupta2018social}. This method improves S-LSTM by introducing a GAN to generate multimodal prediction results.
	\item SoPhie \cite{sadeghian2019sophie}. This method improves GAN-based methods by applying an attention to model social relationships and physical constraints.
	\item P2TIRL \cite{deo2020trajectory}.  A maximum entropy inverse reinforcement learning strategy is introduced to learn a grid-based trajectory prediction method. 
	\item CF-VAE \cite{bhattacharyya2019conditional}. A conditional normalizing flow-based prior is proposed to improve the VAE for the generation of effective predictions.
	\item CGNS \cite{li2019conditional}. Conditional variational divergence minimization and conditional latent space learning are used to predict the future trajectory.
	\item CF-LSTM \cite{xu2020cf}. A feature-cascaded method that can simultaneously capture location and velocity informations is proposed to improve LSTM.
	\item Trajectron++ \cite{salzmann2020trajectron++}. This method presents a modular graph-structured recurrent model that can intergrate agent dynamics and heterogeneous data.
	\item PECNet \cite{mangalam2020not}. This method first predicts distant trajectory endpoints and then generates the future predictions conditioned on endpoints.
	\item TNT \cite{zhao2020tnt}. This method improves PECNet using a scoring and selection stage that ranks and selects multimodal predictions with likelihood scores.
	\item Y-net \cite{mangalam2020goals}. This method improves PECNet and TNT by iteratively predicting intermediate waypoints and trajectories.
\end{itemize}

\begin{table*}[!t]
	\begin{center}
		\resizebox{\textwidth}{13mm}{
			\begin{tabular}{c|c|c|c||c|c|c|c|c|c|c|c|c|c|c|c}
				\hline
				\multicolumn{4}{c||}{Modules} & \multicolumn{2}{c|}{ETH} &  \multicolumn{2}{c|}{HOTEL}  & \multicolumn{2}{c|}{UNIV}  & \multicolumn{2}{c|}{ZARA1}  & \multicolumn{2}{c|}{ZARA2}  & \multicolumn{2}{c}{AVG} \\
				\hline
				$BL$ & $CCM$ & $SCM$ & $SRM$ &  ADE&FDE & ADE&FDE & ADE&FDE & ADE&FDE & ADE&FDE & ADE&FDE \\
				\hline
				\checkmark & & & & 0.48 & 0.94 & 0.19 & 0.35 & 0.30 & 0.61 & 0.22 & 0.42 & 0.16 & 0.32 & 0.27 & 0.53\\ 
				\hline
				& \checkmark & & & 0.33 & 0.57 & 0.08 & 0.10 & 0.28 & 0.36 & 0.09 & 0.10 & 0.07 & 0.09 & 0.17 & 0.24 \\ 
				\hline
				& \checkmark & & \checkmark & 0.28 & 0.53 & 0.07 & 0.08 & 0.24 & 0.35 & 0.07 & 0.09 & 0.05 & 0.09 & 0.14 & 0.23\\
				\hline
				& & \checkmark & \checkmark & 0.19  & 0.35 & 0.06 & 0.07 & 0.13 & 0.21 & 0.06 & 0.07 & 0.05 & 0.08 & 0.10 & 0.16\\
				\hline
		\end{tabular}}
	\end{center}
	\caption{Ablation study of different modules of the proposed method on the ETH/UCY dataset. $BL$ represents the introduced baseline method. $CCM$ denotes the proposed cascaded CVAE module. $SCM$ denotes the proposed slide CVAE module. $SRM$ denotes the proposed socially-aware regression module.}
	\label{tab.4}
\end{table*}

\begin{table}[!t]
	\begin{center}
		\begin{tabular}{c|c|c|c||c|c}
			\hline
			\multicolumn{4}{c||}{Modules} & \multicolumn{2}{c}{Performance} \\
			\hline
			$BL$ & $CCM$ & $SCM$ & $SRM$ &  ADE & FDE \\
			\hline
			\checkmark & & & & 11.56 & 15.71\\ 
			\hline
			& \checkmark & & & 6.12 & 7.87\\ 
			\hline
			& \checkmark & & \checkmark & 4.87 & 6.32\\
			\hline
			& & \checkmark & \checkmark & 2.77 & 3.45\\
			\hline
		\end{tabular}
	\end{center}
	\caption{Ablation study of different modules of the proposed method on the SDD dataset. $BL$ represents the introduced baseline method. $CCM$ denotes the proposed cascaded CVAE module. $SCM$ denotes the proposed slide CVAE module. $SRM$ denotes the proposed socially-aware regression module.}
	\label{tab.5}
\end{table}

\subsection{Quantitative Analysis}
The proposed CSR model was compared with the aforementioned baseline methods on the SDD and ETH/UCY datasets via the ADE and FDE metrics.

\subsubsection{Comparison with VAE-based Methods}

The VAE is widely used to generate the latent variables of trajectory distributions, and many VAE-based prediction methods have been proposed. Unlike these methods, the proposed method uses multiple CVAEs to predict the future points, in a cascaded manner. The proposed CSR was compared with several VAE-based methods, including DESIRE, CF-VAE, and CGNS. As reported in Table \ref{tab.2}, on the SDD, CSR respectively surpassed CF-VAE and DESIRE by 61.3\%/71.7\% and 56.5\%/60.6\% in terms of the ADE/FDE metrics. Moreover, as reported in Table \ref{tab.3}, CSR outperformed CGNS by 71.4\%/76.3\% on the ETH/UCY. This comparison validates that the proposed CSR, which utilizes CVAEs in a cascaded manner, performs better than VAE-based methods.

\subsubsection{Comparison with RNN-based Methods}
The proposed CSR predicts the future trajectory in a sequential manner, which is similar to RNN-based methods. However, RNN based methods do not explicitly encode the predicted points as the past trajectories. To vallidate the updated past trajectory used in the proposed model, it was compared with several RNN-based methods, including DESIRE, CF-VAE, CGNS and Trajectron++. As shown in Table \ref{tab.2}, on the SDD, the proposed CSR exhibited respective improvements over DESIRE and CF-VAE of 56.5\%/60.6\% and 61.3\%/71.7\% in terms of the ADE/FDE metrics. Moreover, as shown in Table \ref{tab.3}, the proposed CSR exhibited respective improvements over CGNS and Trajectron++ of 71.4\%/76.3\% and 26.3\%/52.1\% on the ETH/UCY dataset. This comparison verifies that the proposed method, which explicitly encodes the predicted points as the past trajectories, achieves superior performance as compared to RNN-based methods.

\subsubsection{Comparison with Target-conditioned Methods}
Target-conditioned trajectory prediction methods have recently become popular. These methods first predict the target point and then generate future predictions condtioned on the predicted target. The performance of this type of methods is easily affected by the predicted target; however, target prediction is inherently a difficult problem.  The proposed CSR was compared with three target-conditioned methods, namely PECNet, TNT, and Y-net. As presented in Table \ref{tab.2}, CSR exhibited respective ADE/FDE improvements over TNT, PECNet, and Y-net (best of 5 samples) of 34.5\%/54.6\%, 31.5\%/36.5\%, and 27.1\%/33.6\% on the SDD. Moreover, as illustrated in Table \ref{tab.3}, the proposed CSR respectively outperformed PECNet and Y-net by 51.7\%/52.1\% and 22.2\%/14.8\% on the ETH/UCY dataset. This comparison demonstrates that the proposed CSR performs better than target-conditioned trajectory prediction methods. 

\begin{figure}[!t]
	\centering
	\subfloat[]{
		\begin{minipage}[b]{0.22\textwidth}
			\includegraphics[width=1\textwidth, trim=0 900 500 360, clip]{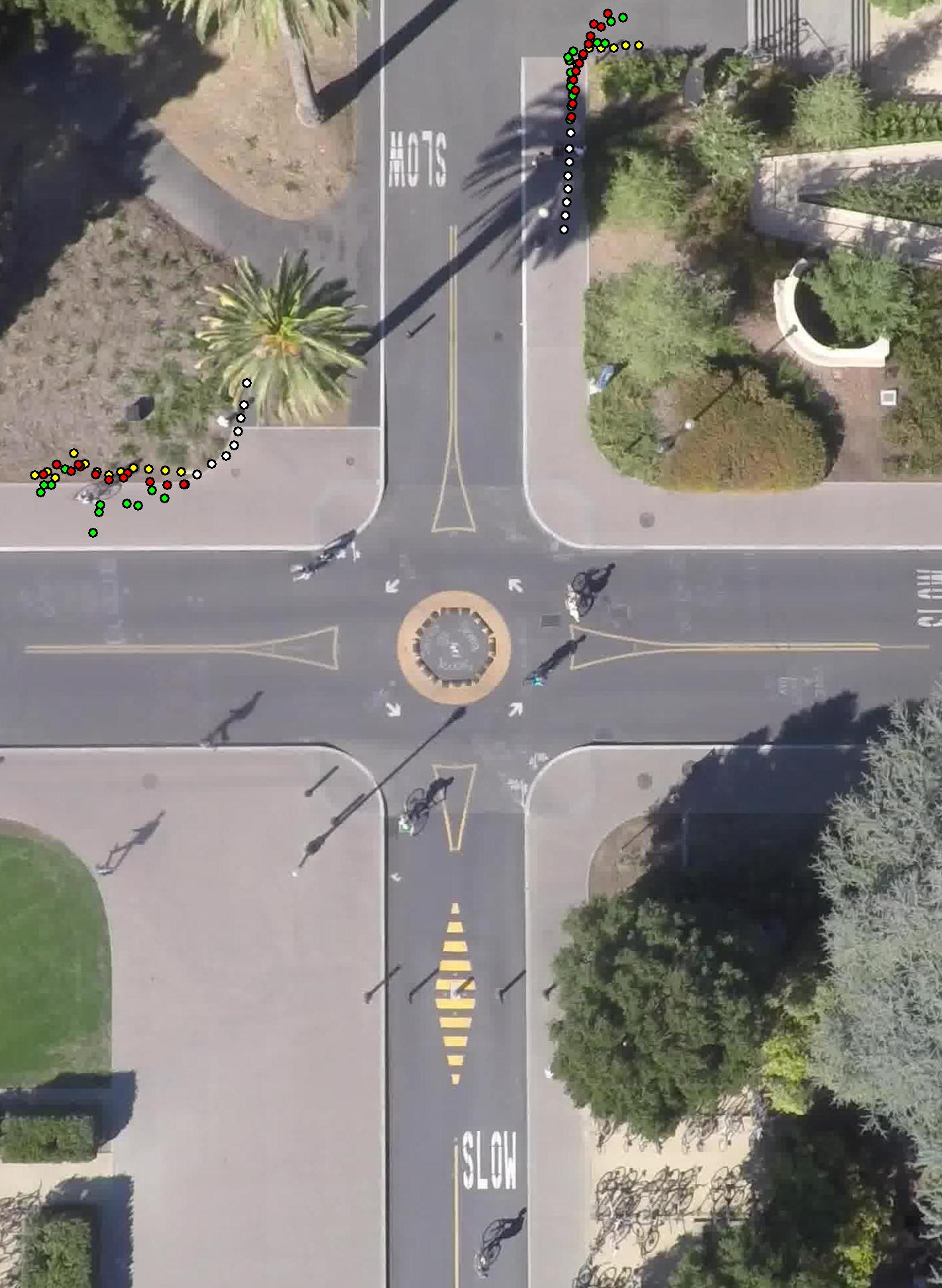} 
		\end{minipage}
		\label{Fig.6a}
	}
	\subfloat[]{
		\begin{minipage}[b]{0.22\textwidth}
			\includegraphics[width=1\textwidth, trim=700 438 400 0, clip]{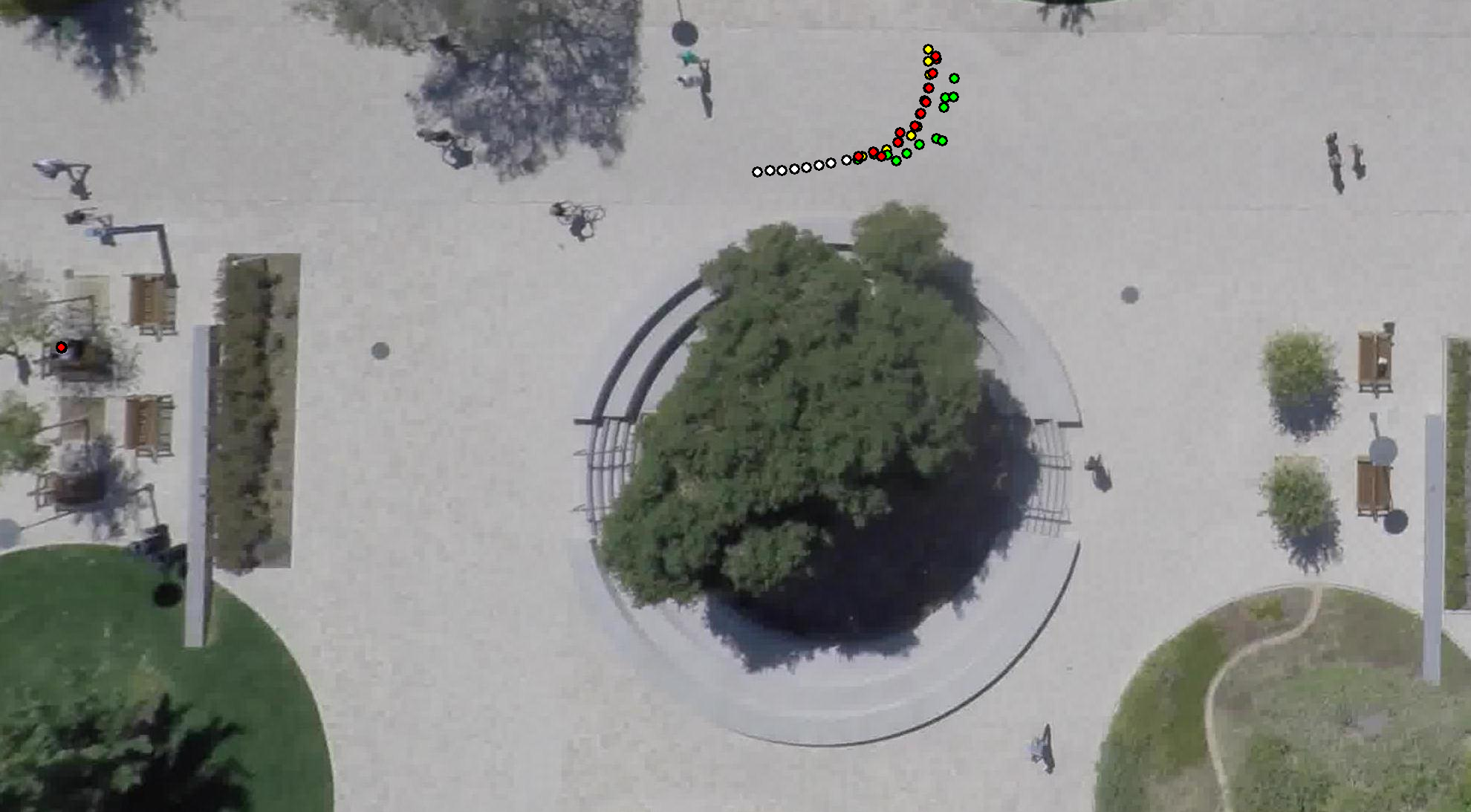} 
		\end{minipage}
		\label{Fig.6b}
	}
	\\
	\subfloat[]{
		\begin{minipage}[b]{0.22\textwidth}
			\includegraphics[width=1\textwidth, trim=100 220 1000 220, clip]{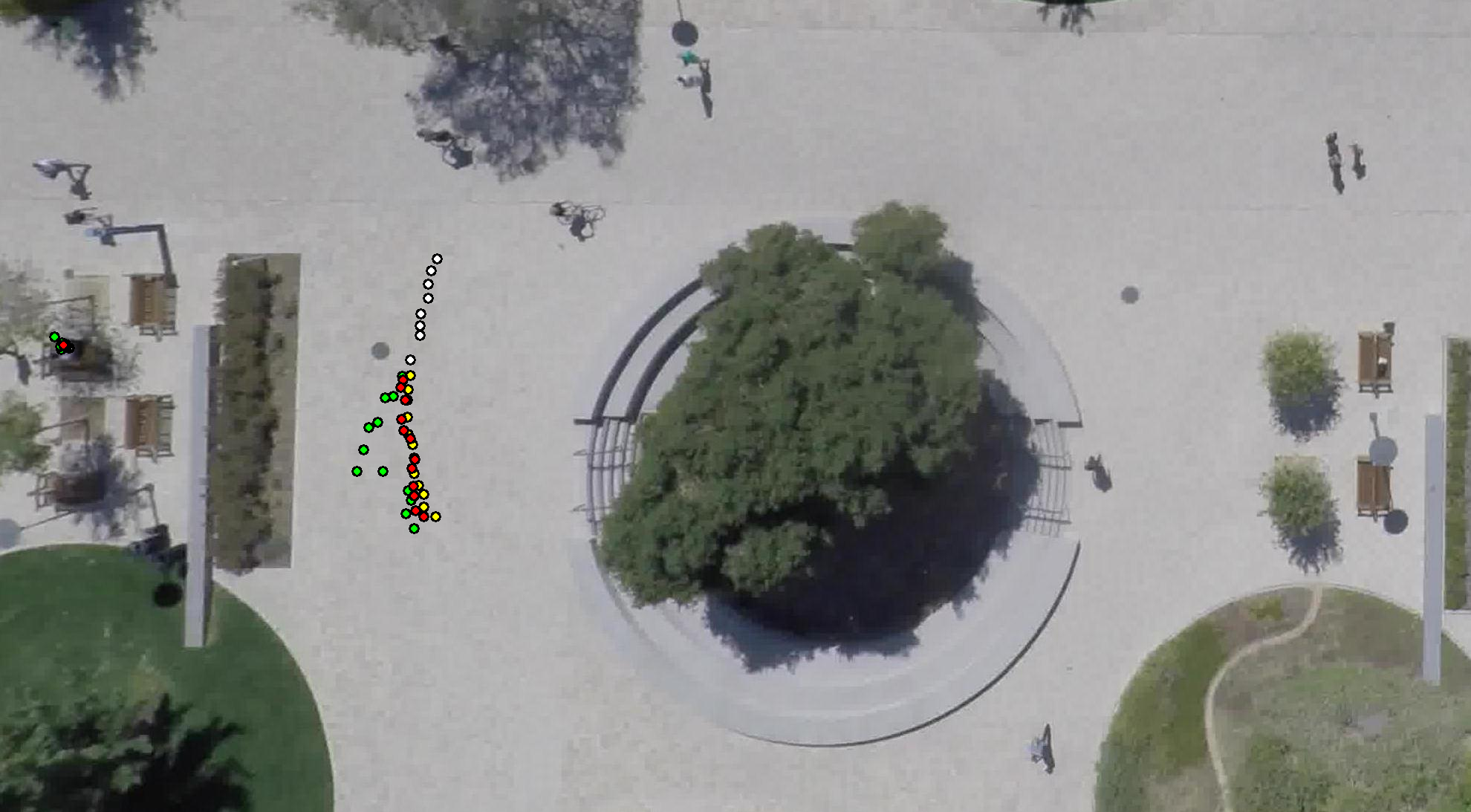}
		\end{minipage}
		\label{Fig.6c}
	}
	\subfloat[]{
		\begin{minipage}[b]{0.22\textwidth}
			\includegraphics[width=1\textwidth, trim=400 468 0 800, clip]{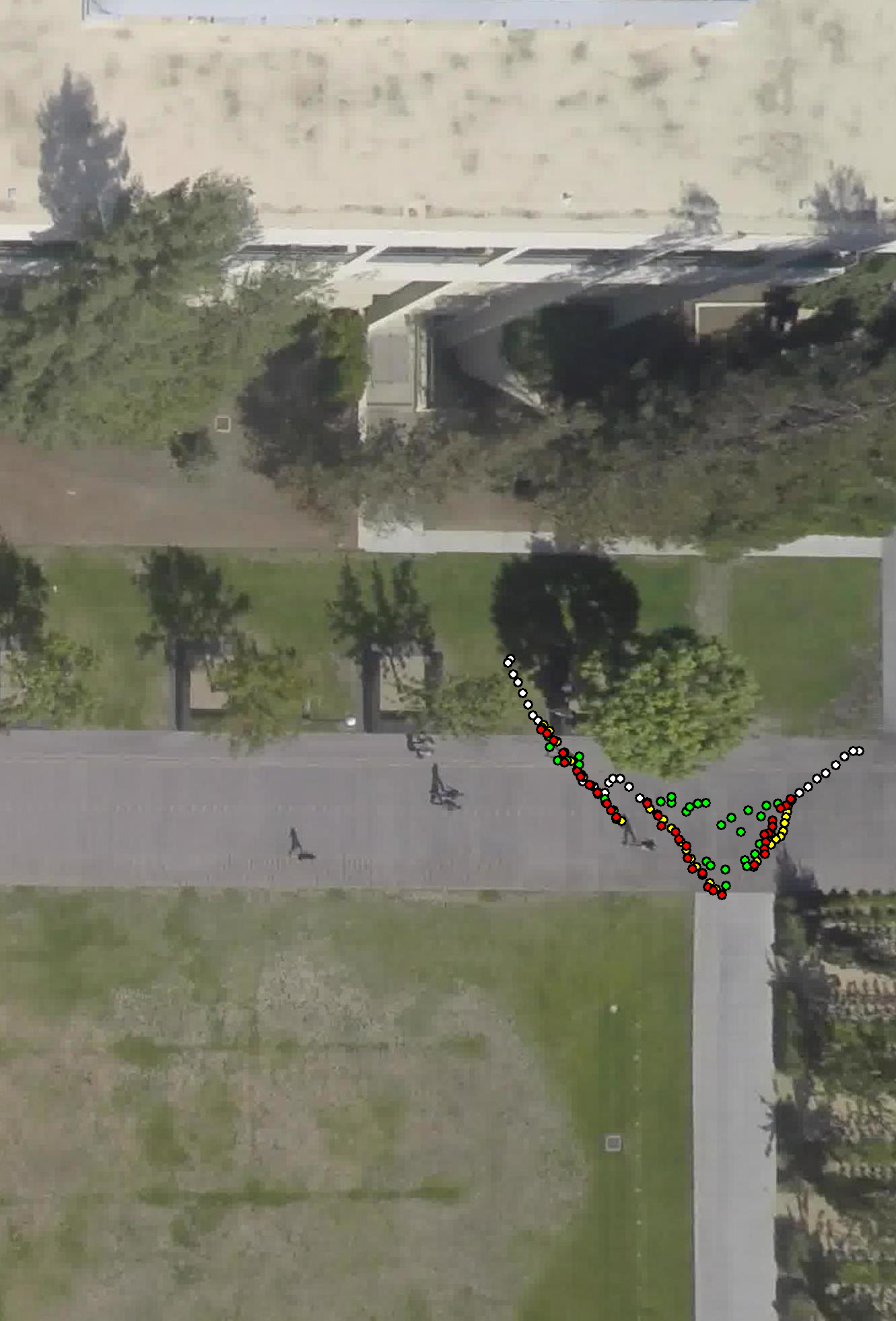}
		\end{minipage}
		\label{Fig.6d}
	}
	\caption{Visualization of the trajectories predicted by the cascaded CVAE module with or without the socially-aware regression module. The visualized trajectories are best predictions sampled from 20 trials. The white dots, yellow dots, green dots, and red dots respectively represent the past trajectories, ground truth future trajectories, predictions without using the socially-aware regression module, and predictions using the socially-aware regression module.}
	\label{Fig.6}
\end{figure}

\subsubsection{Comparison with State-of-the-art Methods}
Tables \ref{tab.2} and \ref{tab.3} reveal that Y-net achieved the best performance among all the selected baselines on the SDD and ETH/UCY. Compared to Y-net, the proposed CSR exhibited improvements in the ADE/FDE metrics of 38.0\%/46.7\% (best of 20 samples) on the SDD and 22.2\%/14.8\% on the ETH/UCY dataset, thereby achieving new state-of-the-art performance.

\begin{figure*}[!t]
	\centering
	\subfloat[]{
		\begin{minipage}[b]{0.232\textwidth}
			\includegraphics[width=1\textwidth, trim=40 1015 40 30, clip]{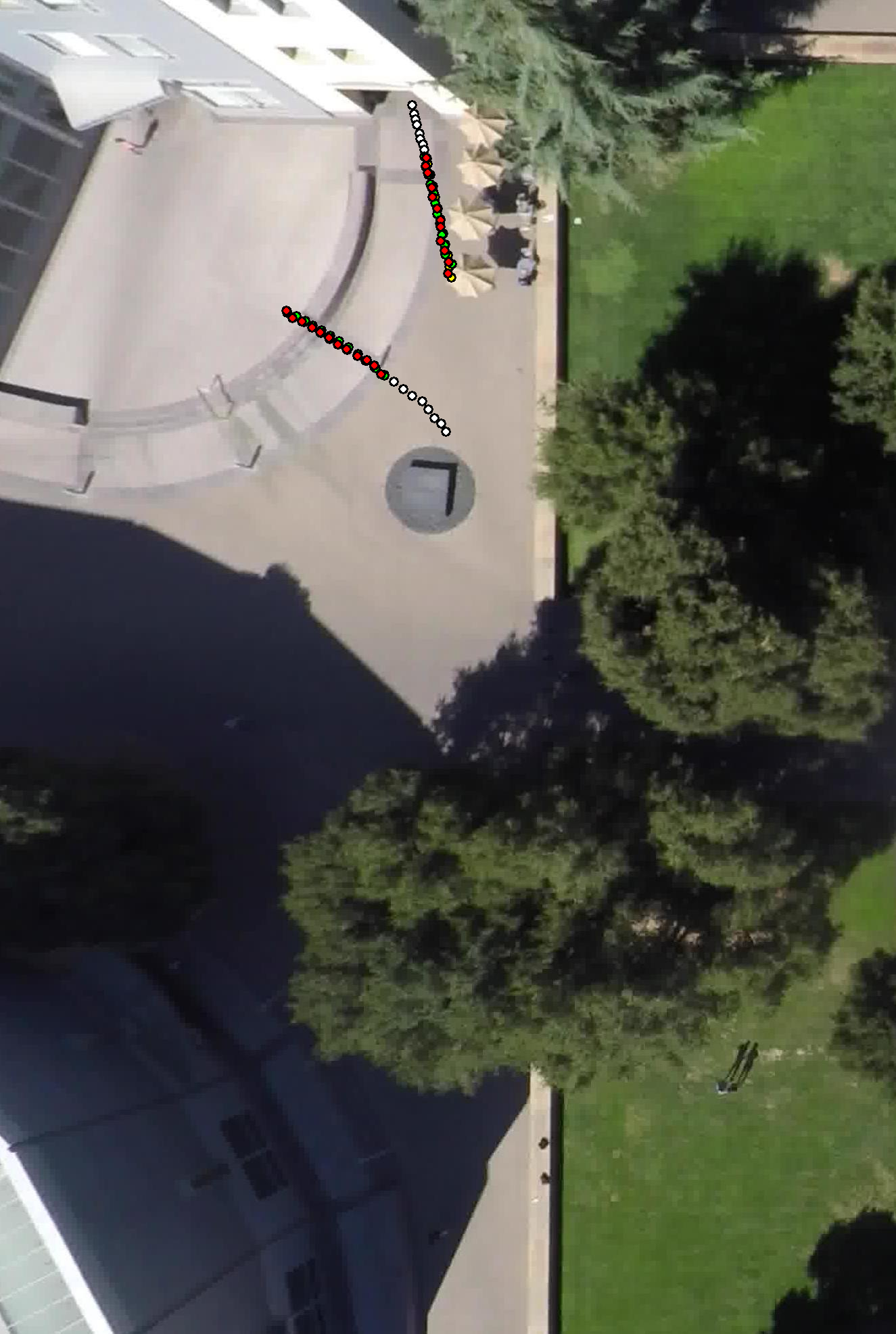} 
		\end{minipage}
		\label{Fig.7a}
	}
	\subfloat[]{
		\begin{minipage}[b]{0.232\textwidth}
			\includegraphics[width=1\textwidth, trim=40 30 40 1000, clip]{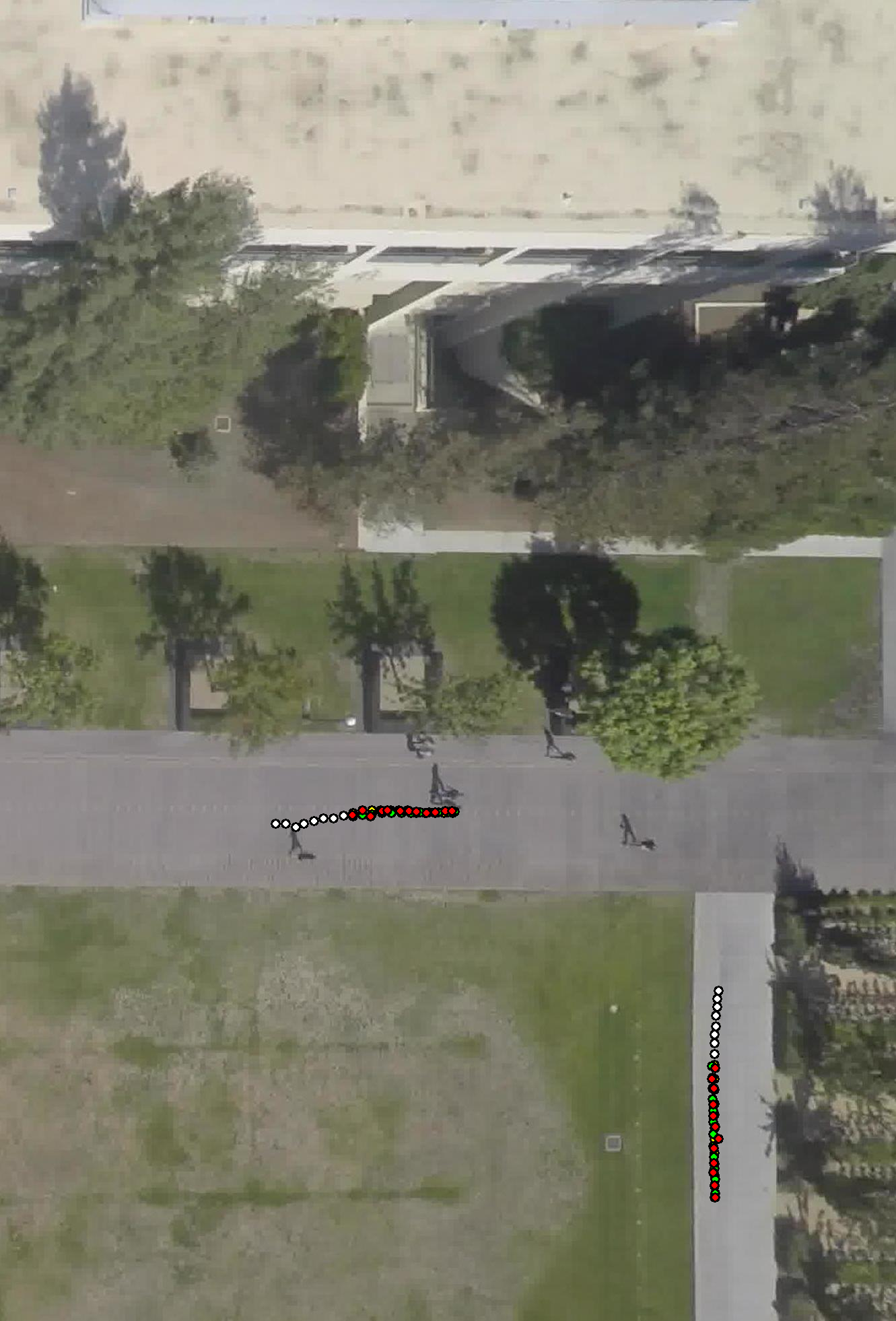} 
		\end{minipage}
		\label{Fig.7b}
	}
	\subfloat[]{
		\begin{minipage}[b]{0.232\textwidth}
			\includegraphics[width=1\textwidth, trim=40 400 40 497, clip]{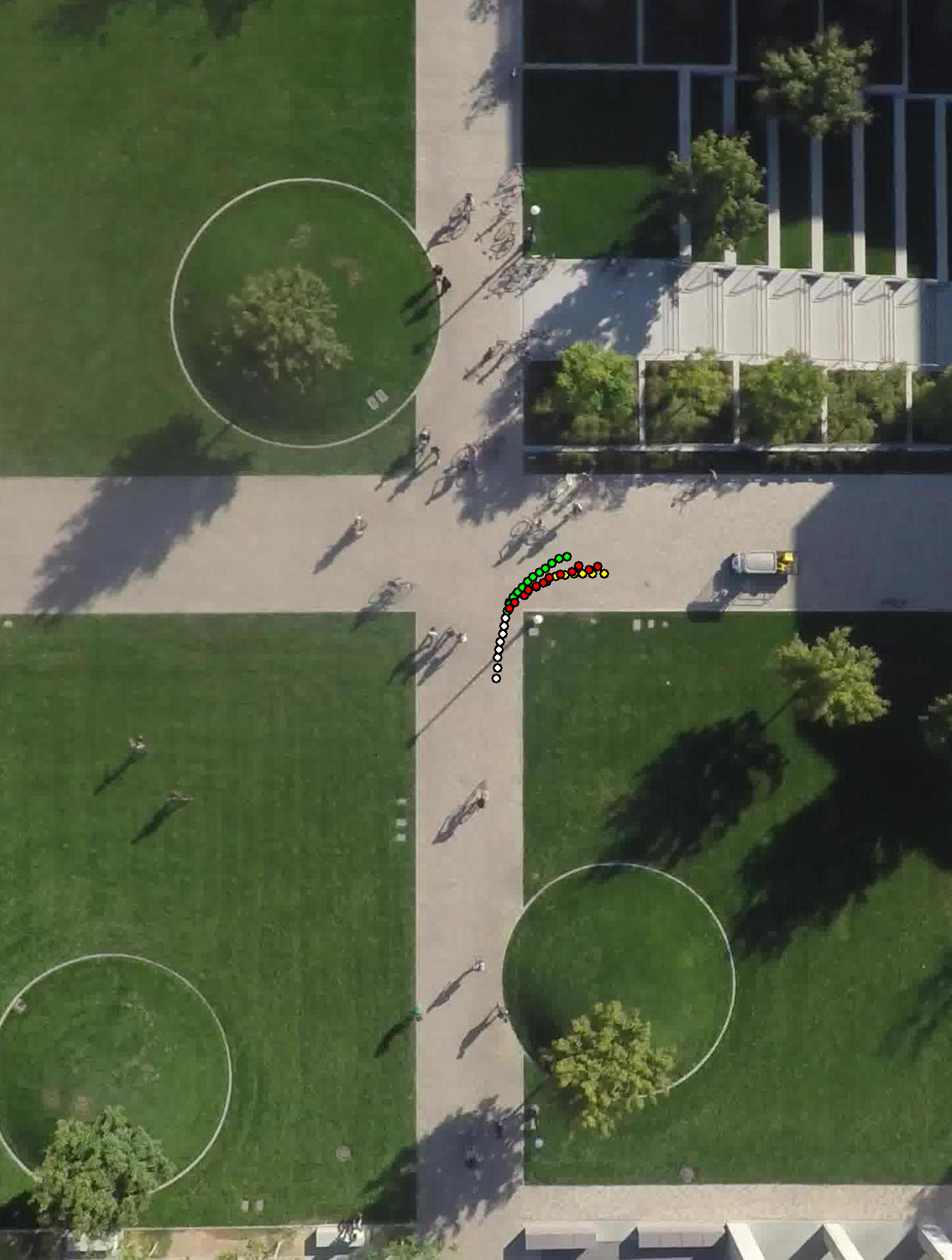}
		\end{minipage}
		\label{Fig.7c}
	}
	\subfloat[]{
		\begin{minipage}[b]{0.232\textwidth}
			\includegraphics[width=1\textwidth, trim=40 40 40 900, clip]{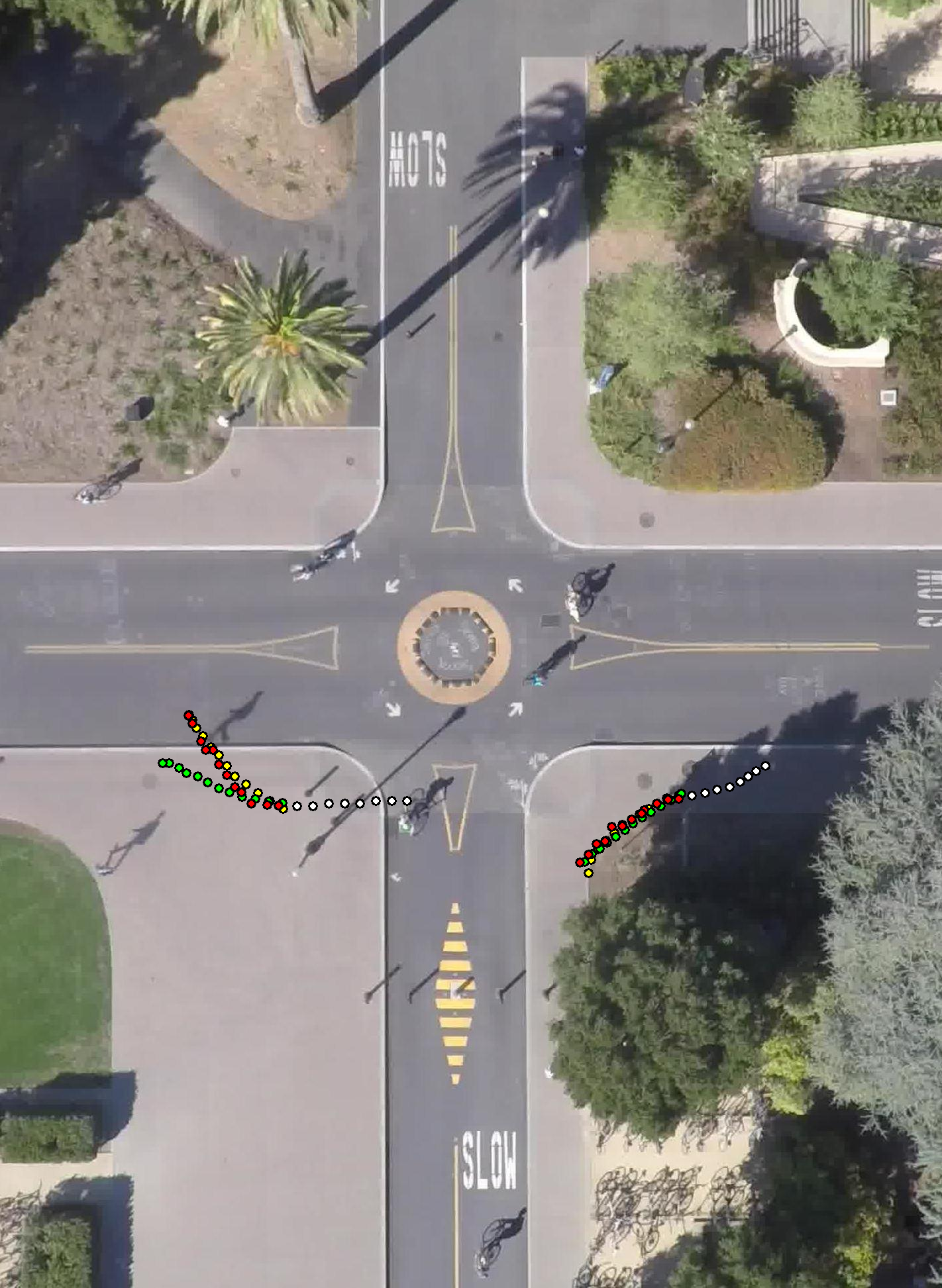}
		\end{minipage}
		\label{Fig.7d}
	}
	\\ 
	\subfloat[]{
		\begin{minipage}[b]{0.232\textwidth}
			\includegraphics[width=1\textwidth, trim=40 1015 40 30, clip]{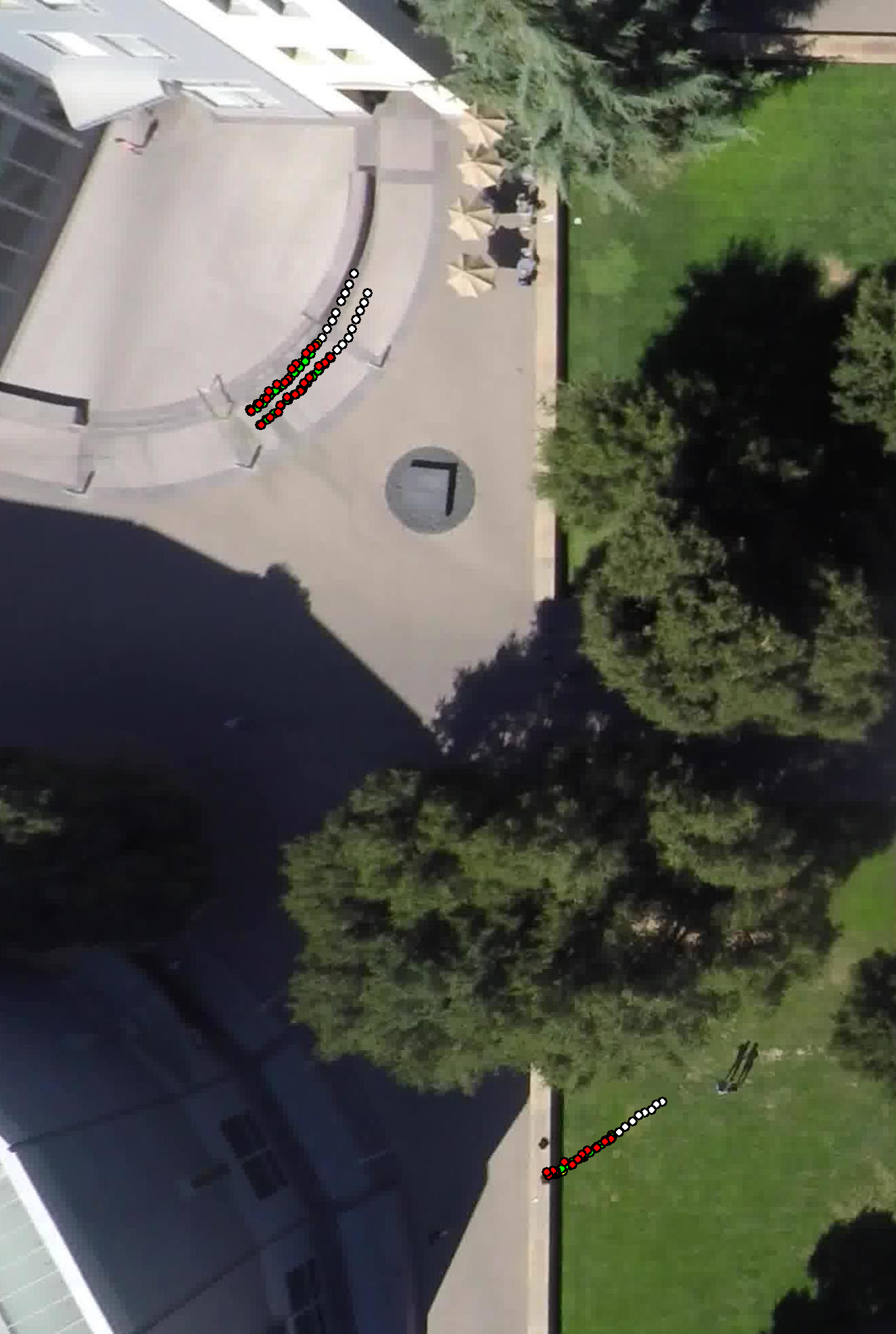} 
		\end{minipage}
		\label{Fig.7e}
	}
	\subfloat[]{
		\begin{minipage}[b]{0.232\textwidth}
			\includegraphics[width=1\textwidth, trim=40 30 40 1000, clip]{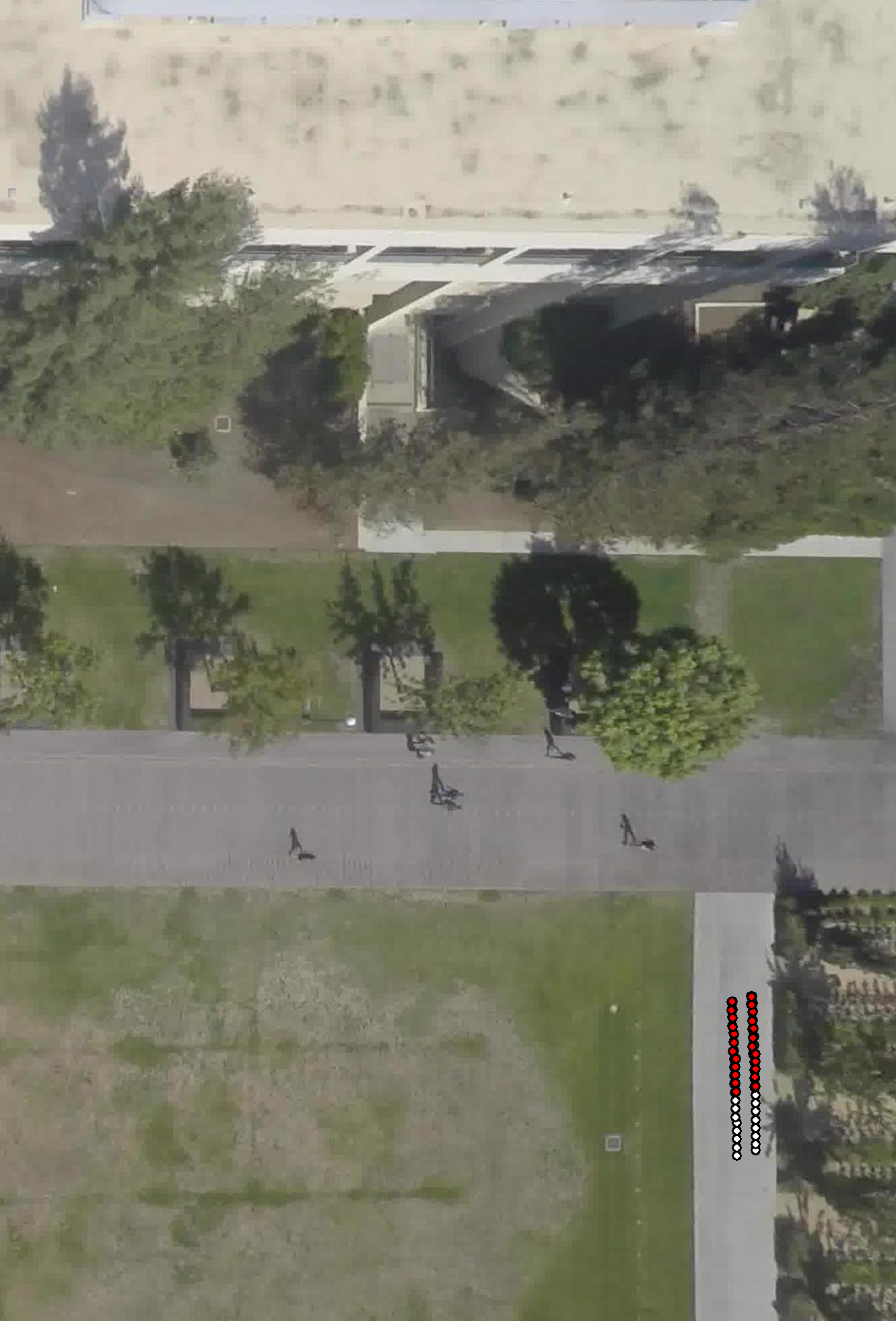}
		\end{minipage}
		\label{Fig.7f}
	}
	\subfloat[]{
		\begin{minipage}[b]{0.232\textwidth}
			\includegraphics[width=1\textwidth, trim=40 857 40 40, clip]{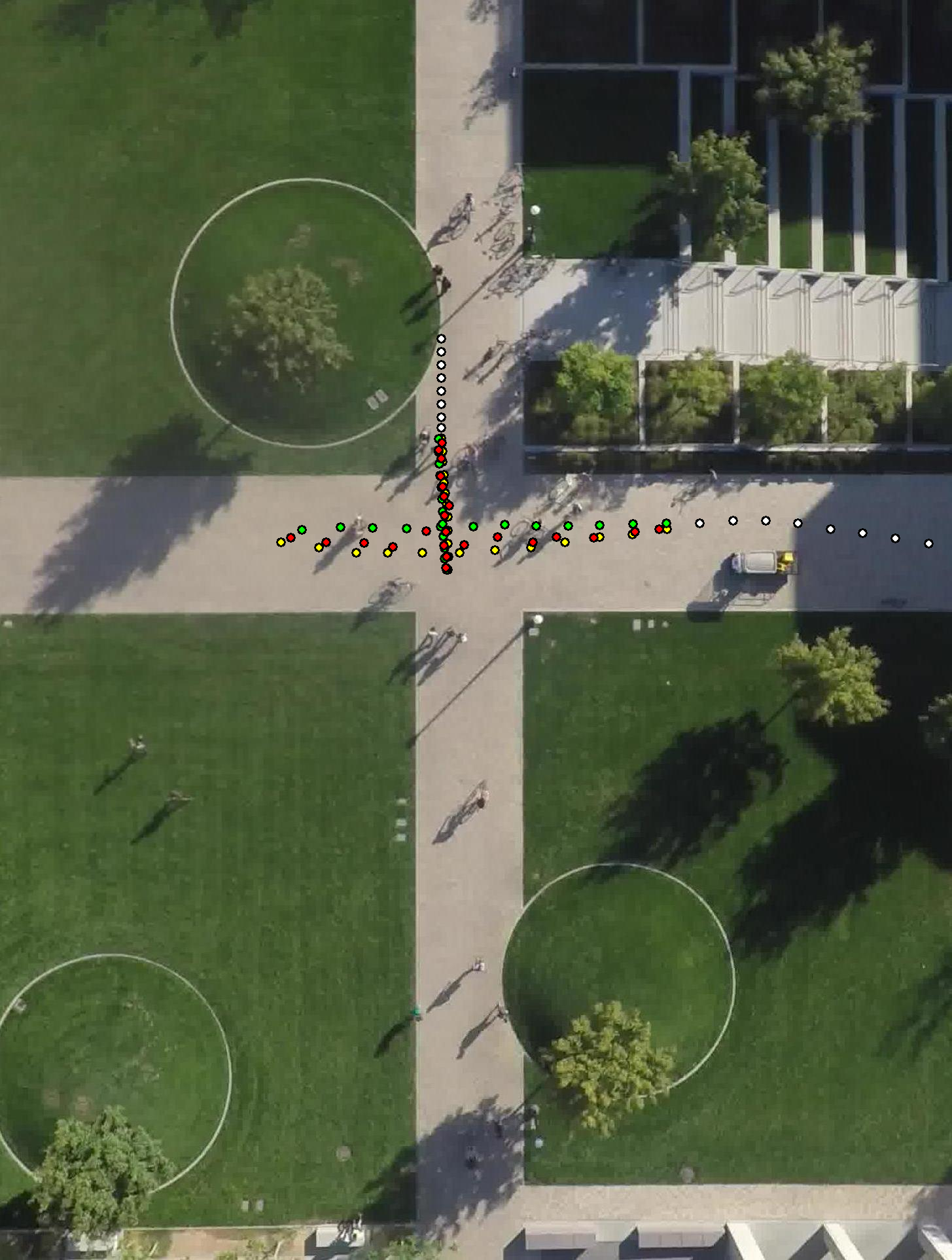}
		\end{minipage}
		\label{Fig.7g}
	}
	\subfloat[]{
		\begin{minipage}[b]{0.232\textwidth}
			\includegraphics[width=1\textwidth, trim=80 40 0 900, clip]{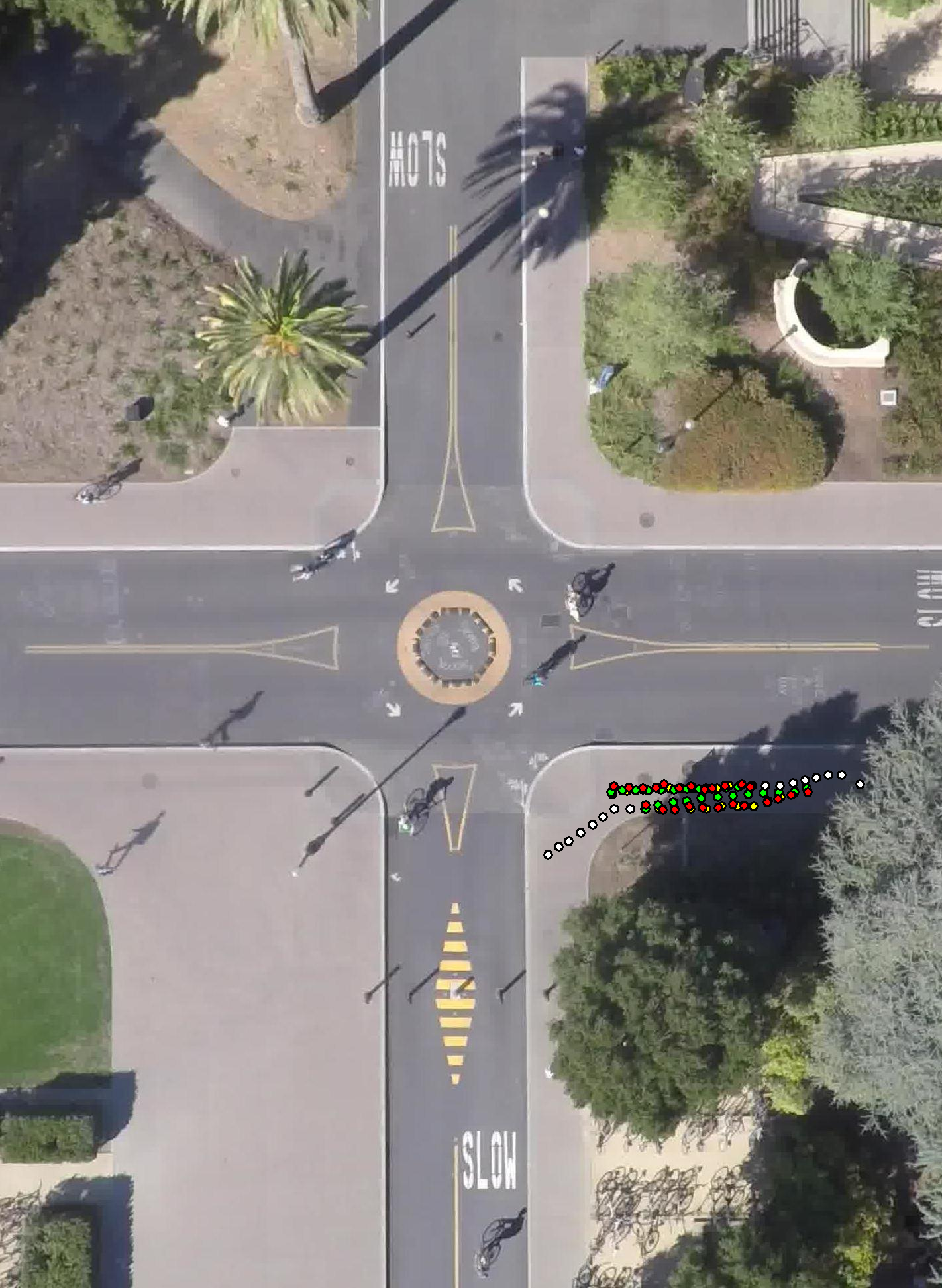}
		\end{minipage}
		\label{Fig.7h}
	}
	\caption{Qualitative results of CSR in SDD dataset. The visualized trajectories are best predictions sampled from 20 trials. The white dots, yellow dots, green dots, and red dots respectively represent the past trajectories, ground truth future trajectories, predictions of PECNet, and predictions of the proposed CSR.}
	\label{Fig.7}
\end{figure*}

\begin{table}[!t]
	\begin{center}
		\begin{tabular}{c|c|c||c|c}
			\hline
			\multicolumn{3}{c||}{Modules} & \multicolumn{2}{c}{Performance} \\
			\hline
			$CCM$ & $SCM$ & $SRM$ &  Parameters & Inference time \\
			\hline
			\checkmark & & \checkmark & 15.88 M & 0.32 S/B\\ 
			\hline
			& \checkmark & \checkmark & 2.24 M & 0.30 S/B\\ 
			\hline
		\end{tabular}
	\end{center}
	\caption{Analysis of the model efficiency of the cascaded CVAE module and the slide CVAE module on the SDD dataset. $CCM$ denotes the proposed cascaded CVAE module. $SCM$ denotes the proposed slide CVAE module. $SRM$ denotes the proposed socially-aware regression module. "S" represents million and "S/B" represents seconds per batch. All experiments are evaluated on an NVIDIA 1080TI GPU.}
	\label{tab.6}
\end{table}

\begin{figure*}[htbp]
	\centering
	\subfloat[Scene: ETH]{
		\begin{minipage}[b]{0.28\textwidth}
			\includegraphics[width=\textwidth, trim=30 30 30 30, clip]{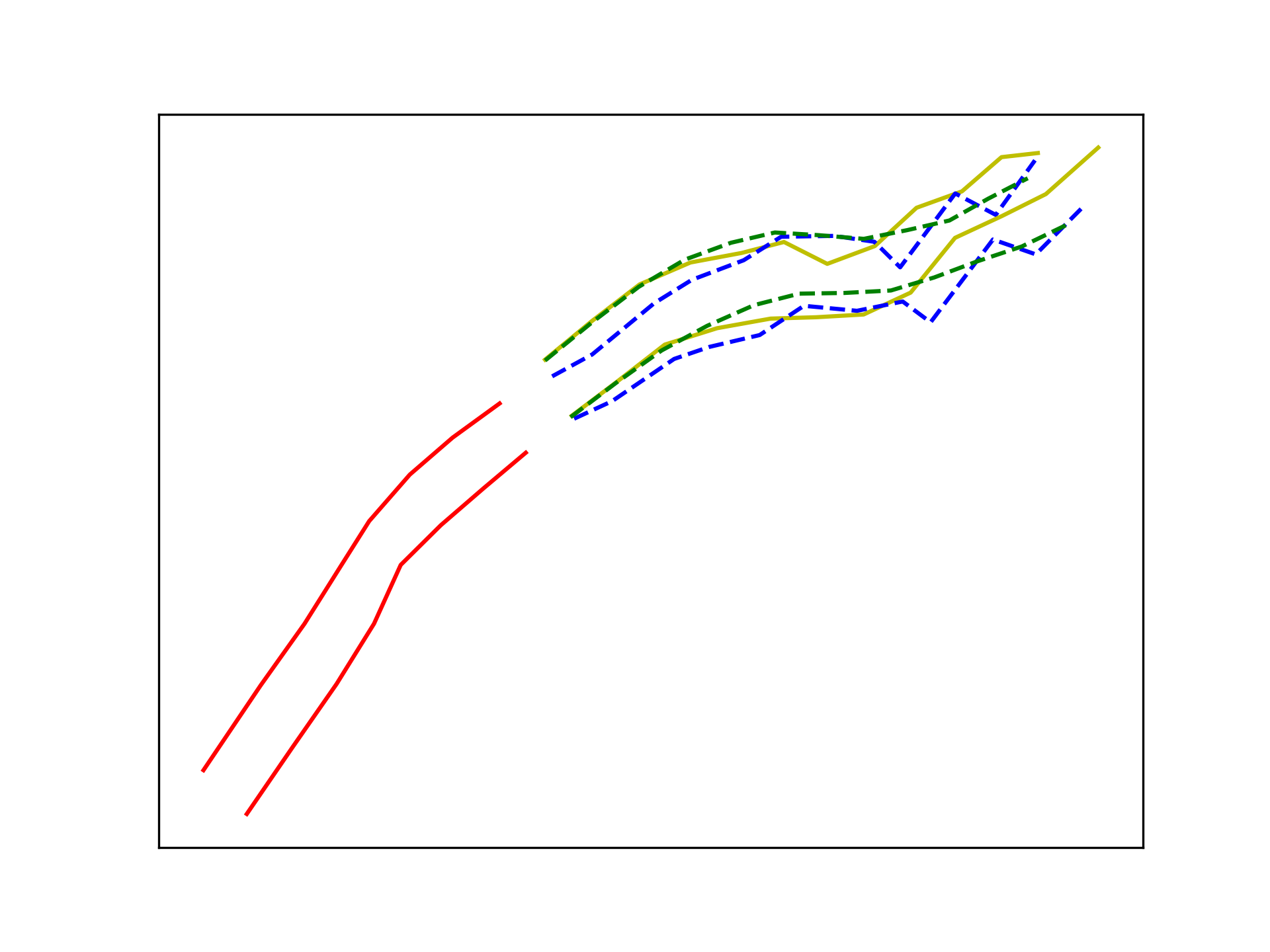}
		\end{minipage}
		\begin{minipage}[b]{0.28\textwidth}
			\includegraphics[width=\textwidth, trim=30 30 30 30, clip]{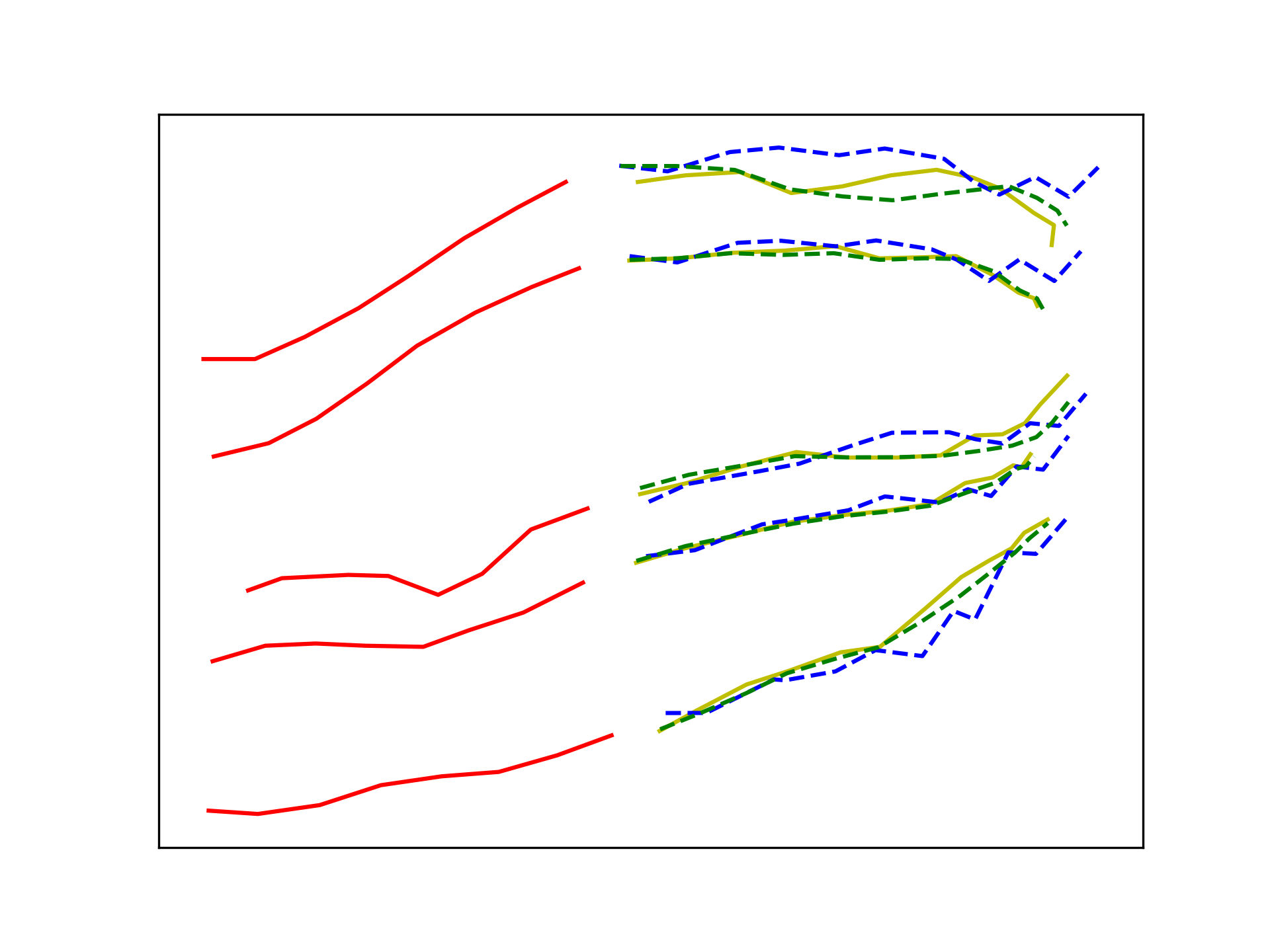}
		\end{minipage}
		\begin{minipage}[b]{0.28\textwidth}
			\includegraphics[width=\textwidth, trim=30 30 30 30, clip]{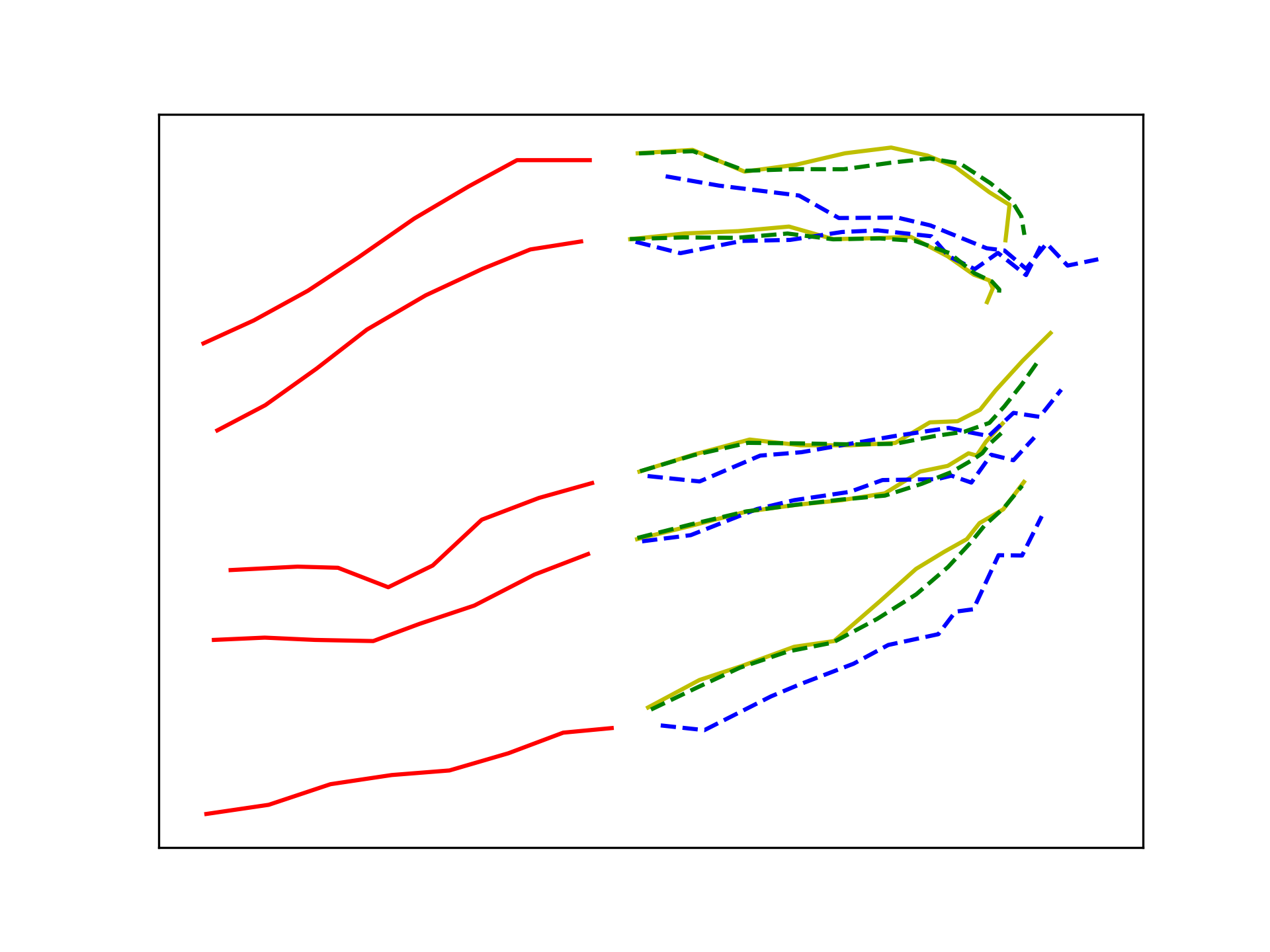}
		\end{minipage}
	}
	\\
	\subfloat[Scene: HOTEL]{
		\begin{minipage}[b]{0.28\textwidth}
			\includegraphics[width=\textwidth, trim=30 30 30 30, clip]{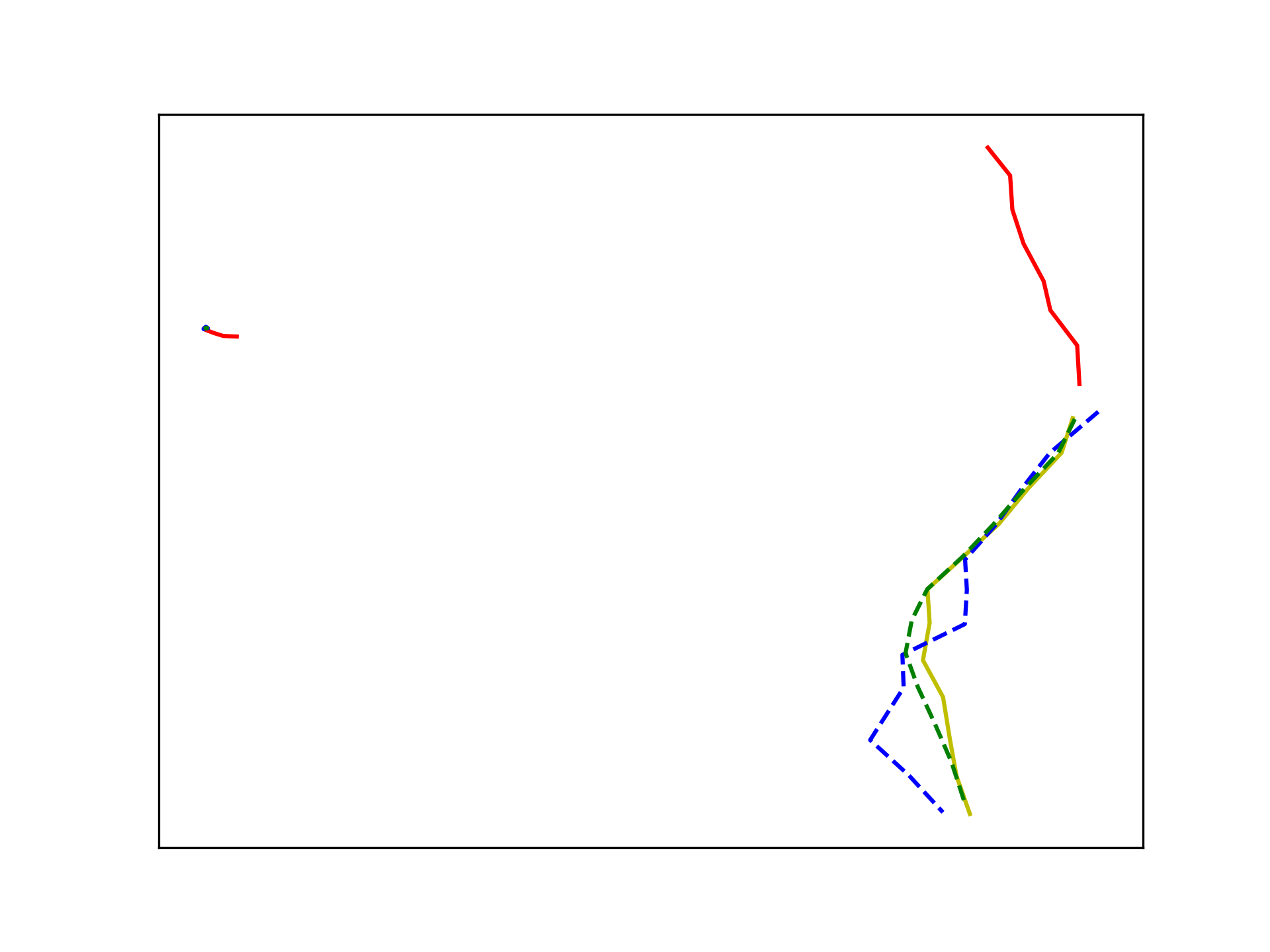}
		\end{minipage}
		\begin{minipage}[b]{0.28\textwidth}
			\includegraphics[width=\textwidth, trim=30 30 30 30, clip]{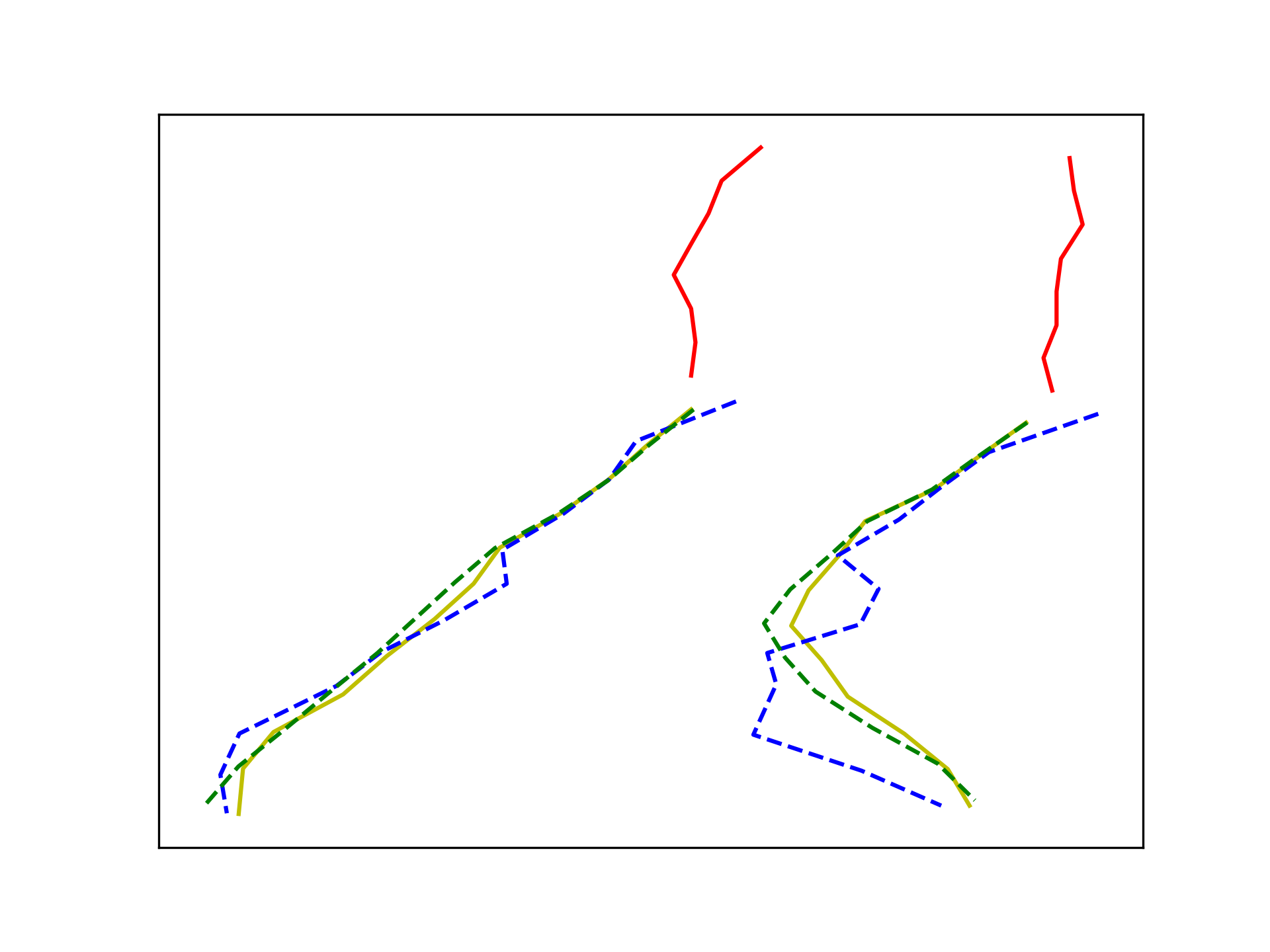}
		\end{minipage}
		\begin{minipage}[b]{0.28\textwidth}
			\includegraphics[width=\textwidth, trim=30 30 30 30, clip]{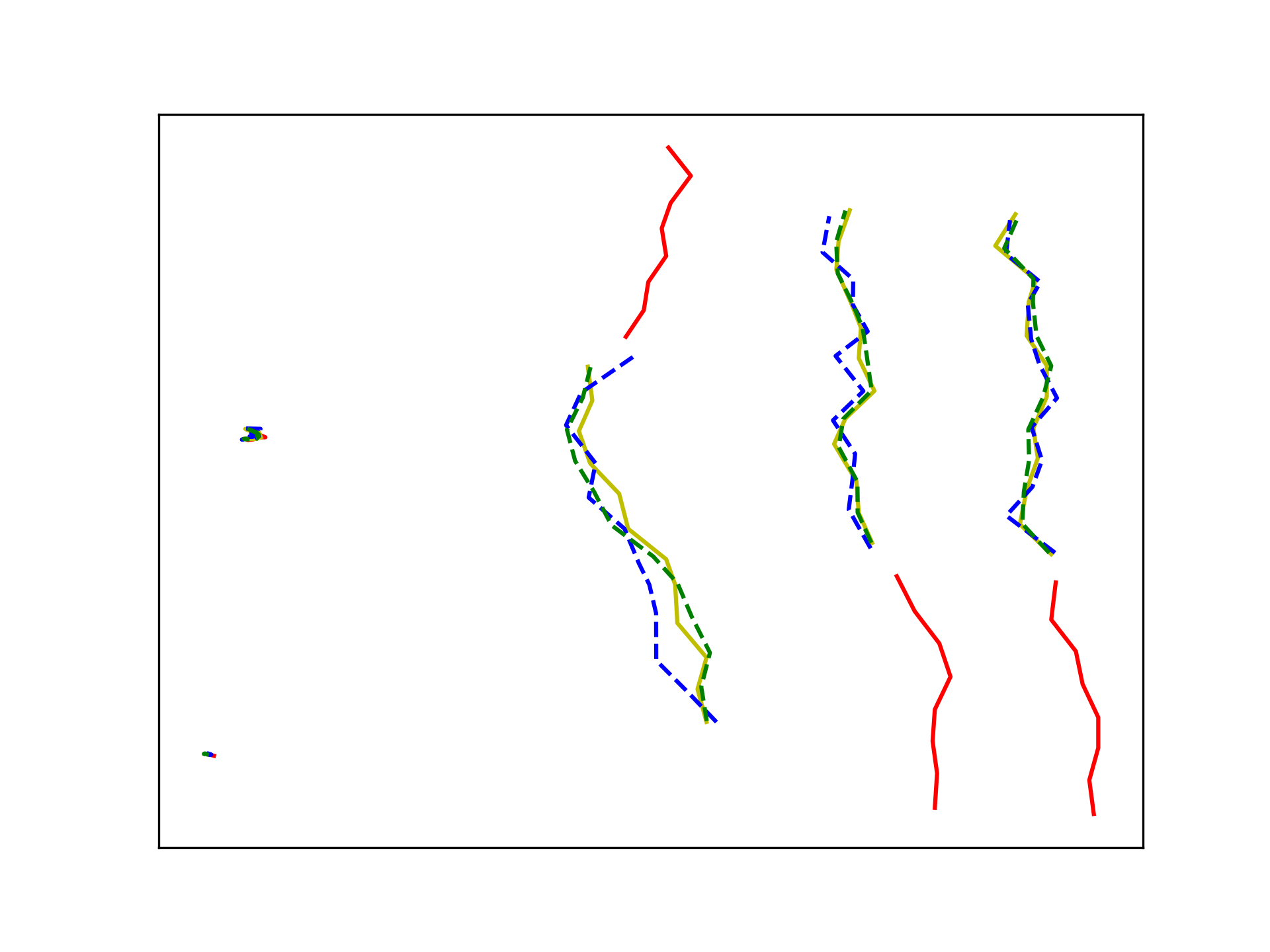}
		\end{minipage}
	}
	\\
	\subfloat[Scene: ZARA1]{
		\begin{minipage}[b]{0.28\textwidth}
			\includegraphics[width=\textwidth, trim=30 30 30 30, clip]{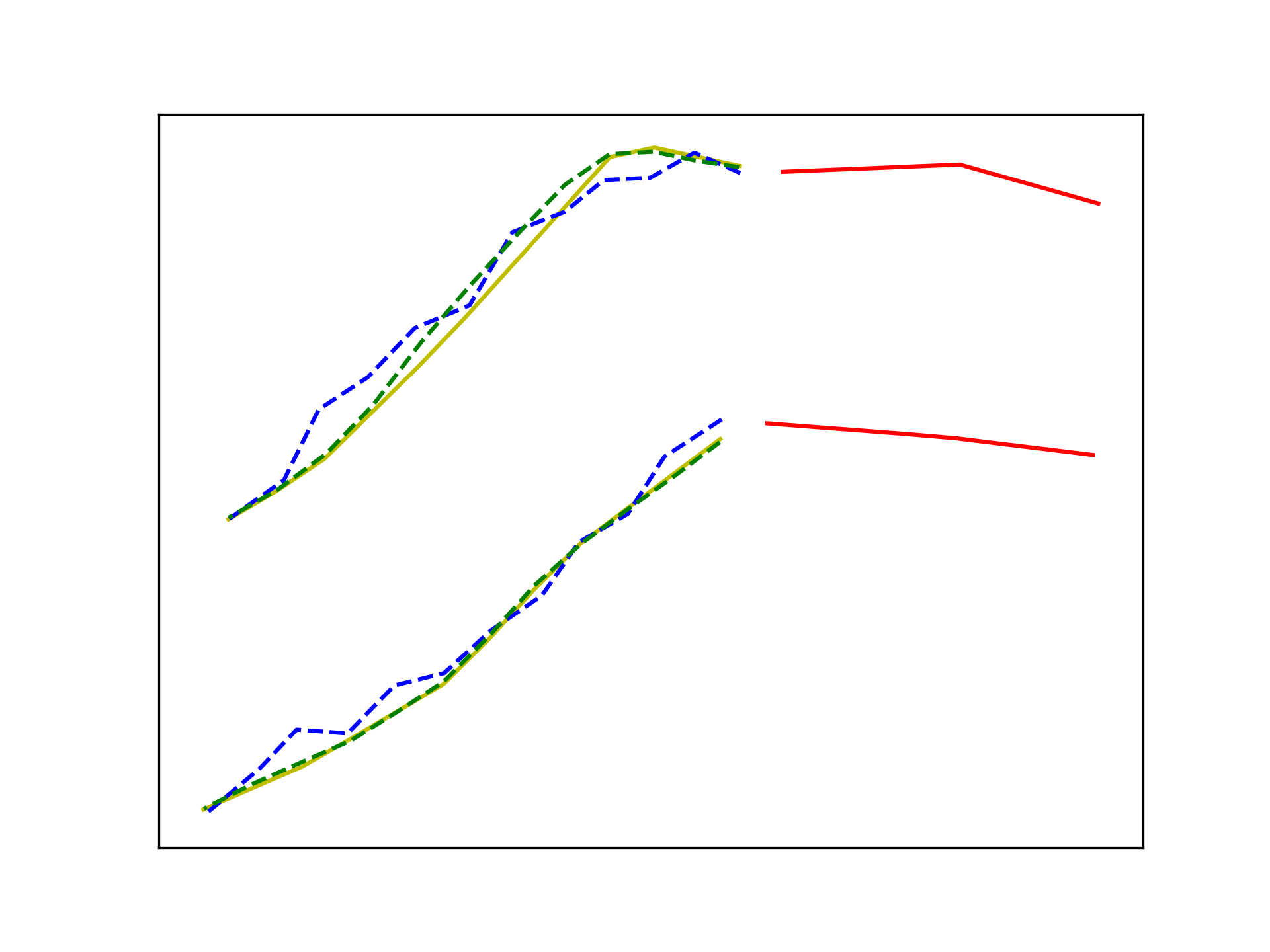}
		\end{minipage}
		\begin{minipage}[b]{0.28\textwidth}
			\includegraphics[width=\textwidth, trim=30 30 30 30, clip]{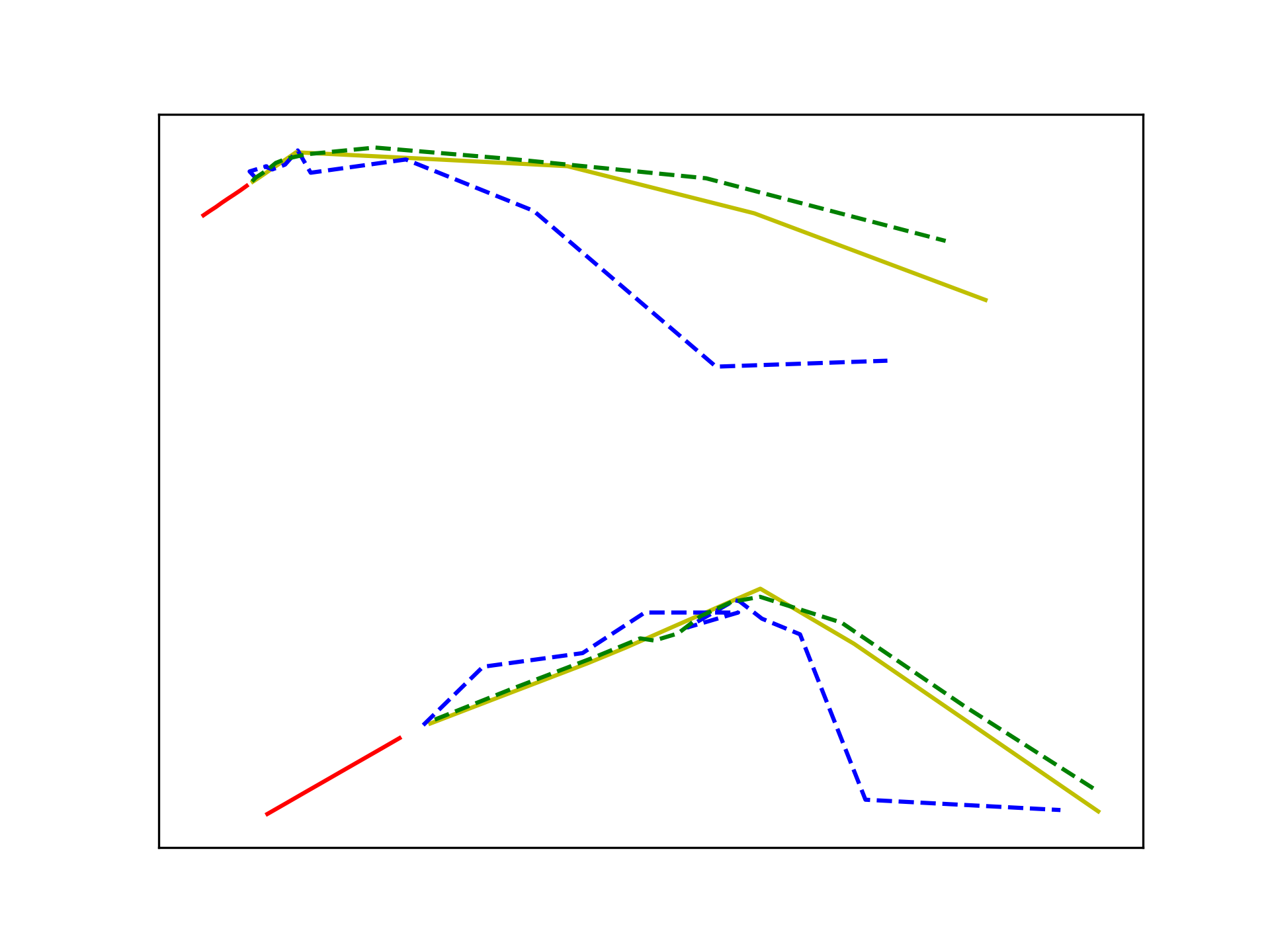}
		\end{minipage}
		\begin{minipage}[b]{0.28\textwidth}
			\includegraphics[width=\textwidth, trim=30 30 30 30, clip]{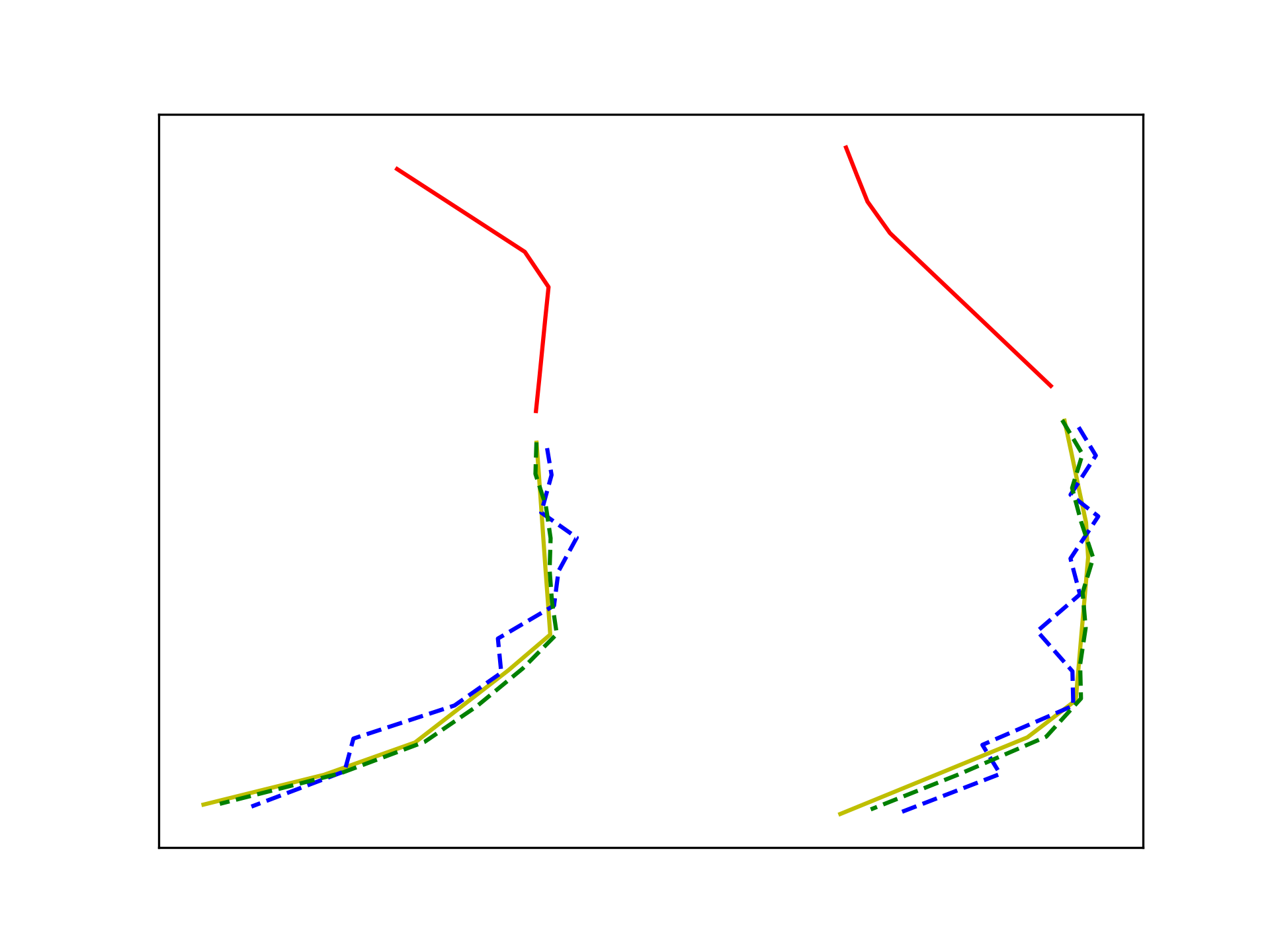}
		\end{minipage}
	}
	\\
	\subfloat[Scene: ZARA2]{
		\begin{minipage}[b]{0.28\textwidth}
			\includegraphics[width=\textwidth, trim=30 30 30 30, clip]{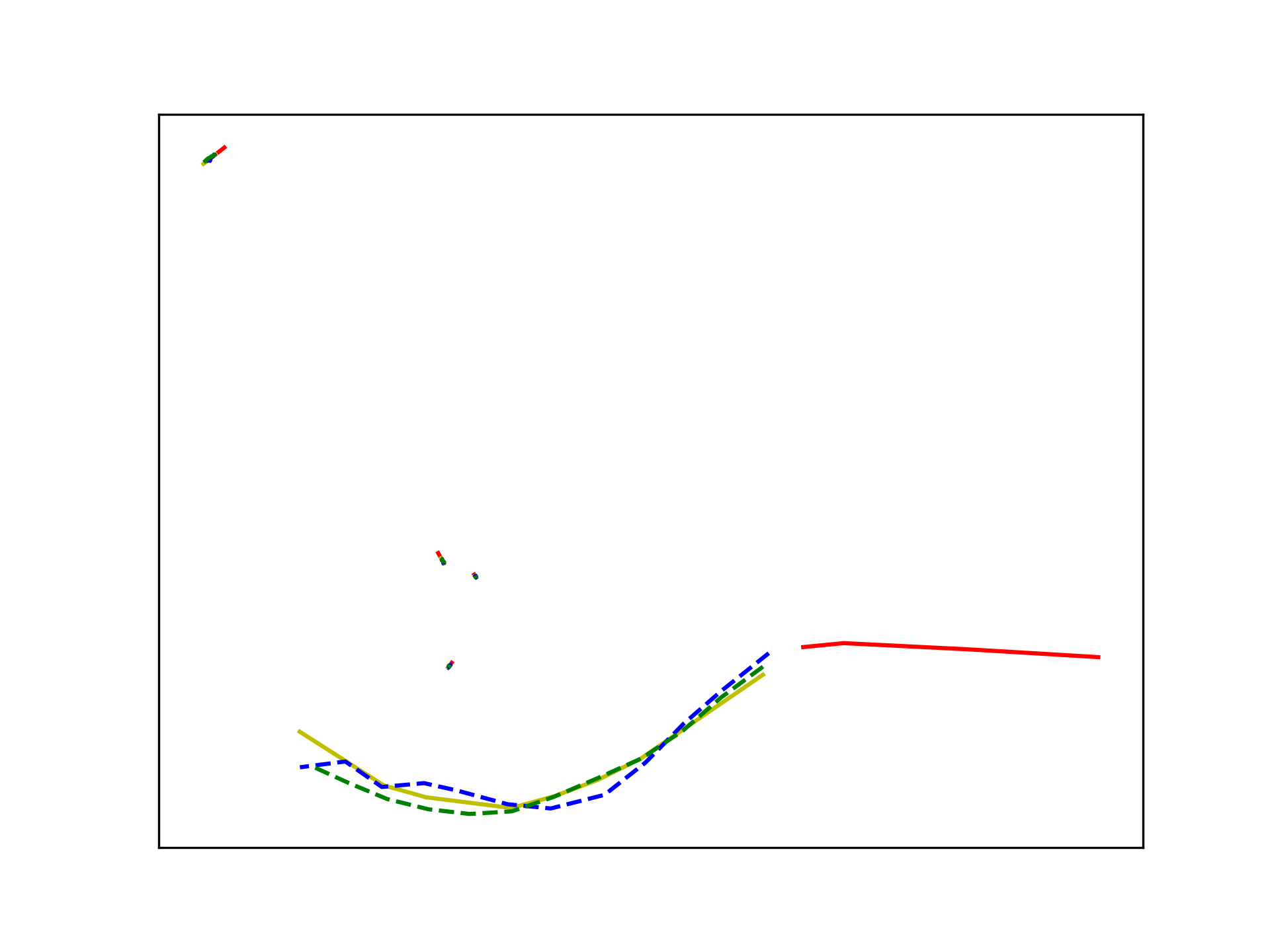}
		\end{minipage}
		\begin{minipage}[b]{0.28\textwidth}
			\includegraphics[width=\textwidth, trim=30 30 30 30, clip]{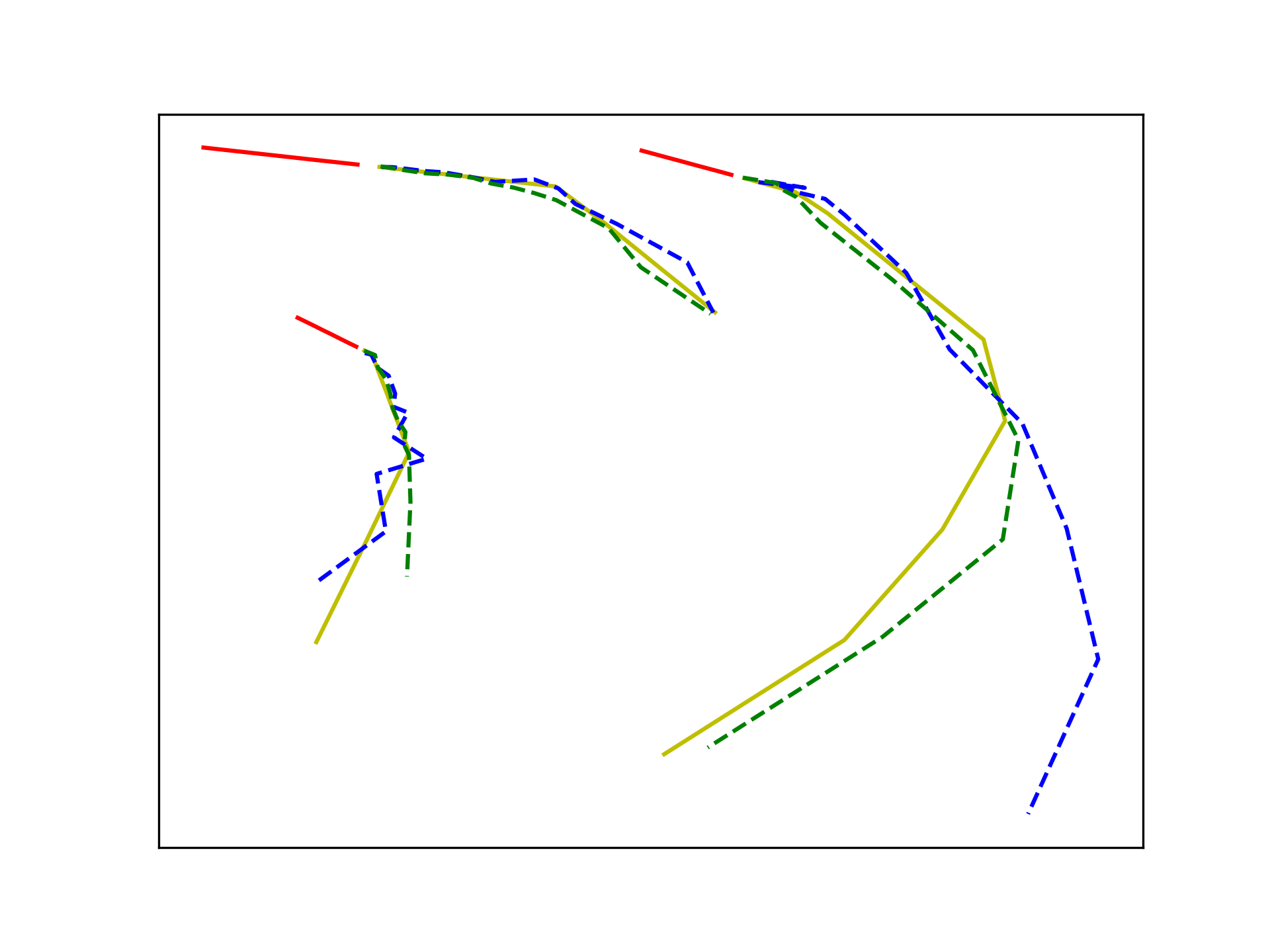}
		\end{minipage}
		\begin{minipage}[b]{0.28\textwidth}
			\includegraphics[width=\textwidth, trim=30 30 30 30, clip]{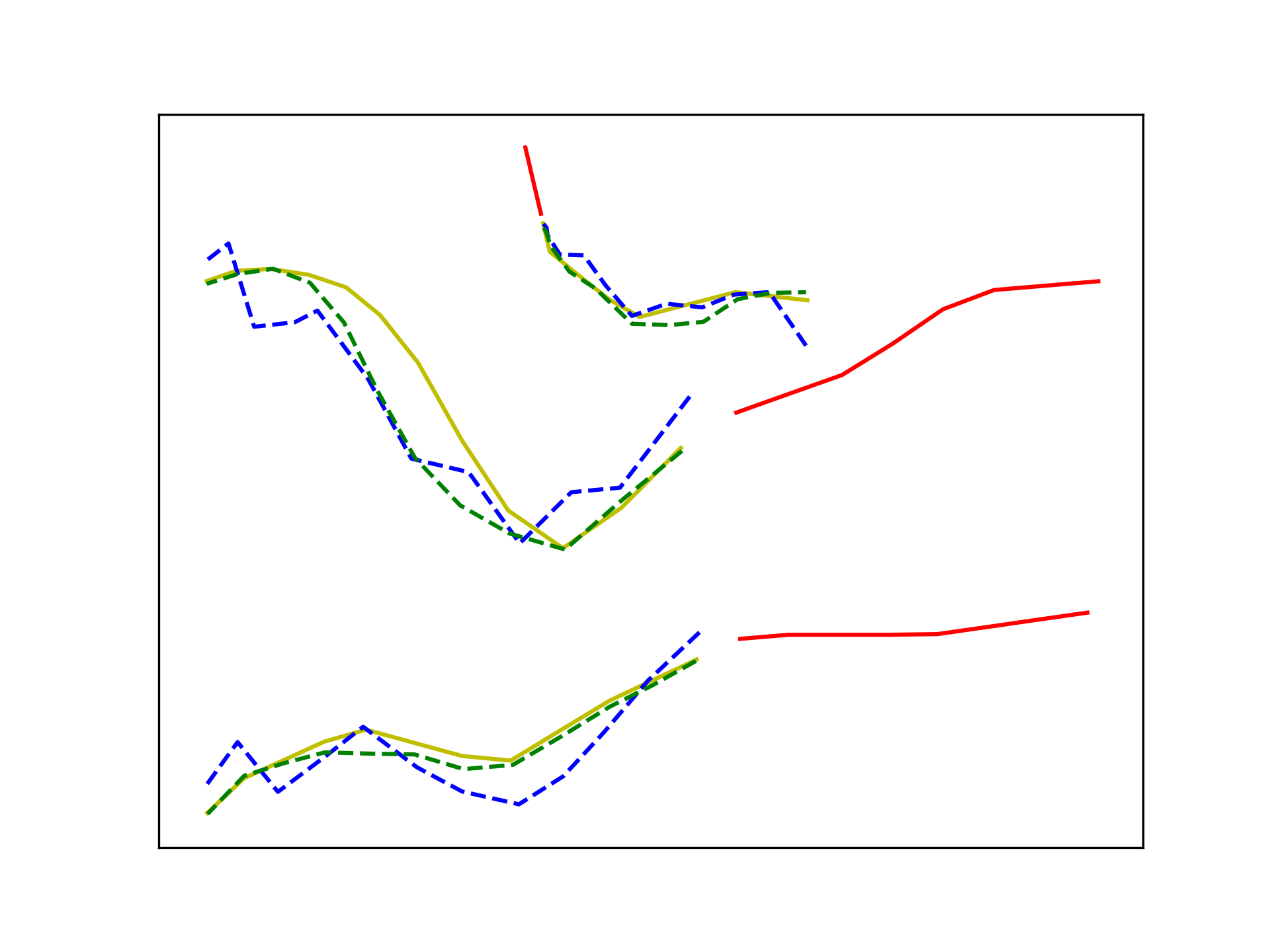}
		\end{minipage}
	}
	\\
	\subfloat[Scene: UNIV]{
		\begin{minipage}[b]{0.28\textwidth}
			\includegraphics[width=\textwidth, trim=30 30 30 30, clip]{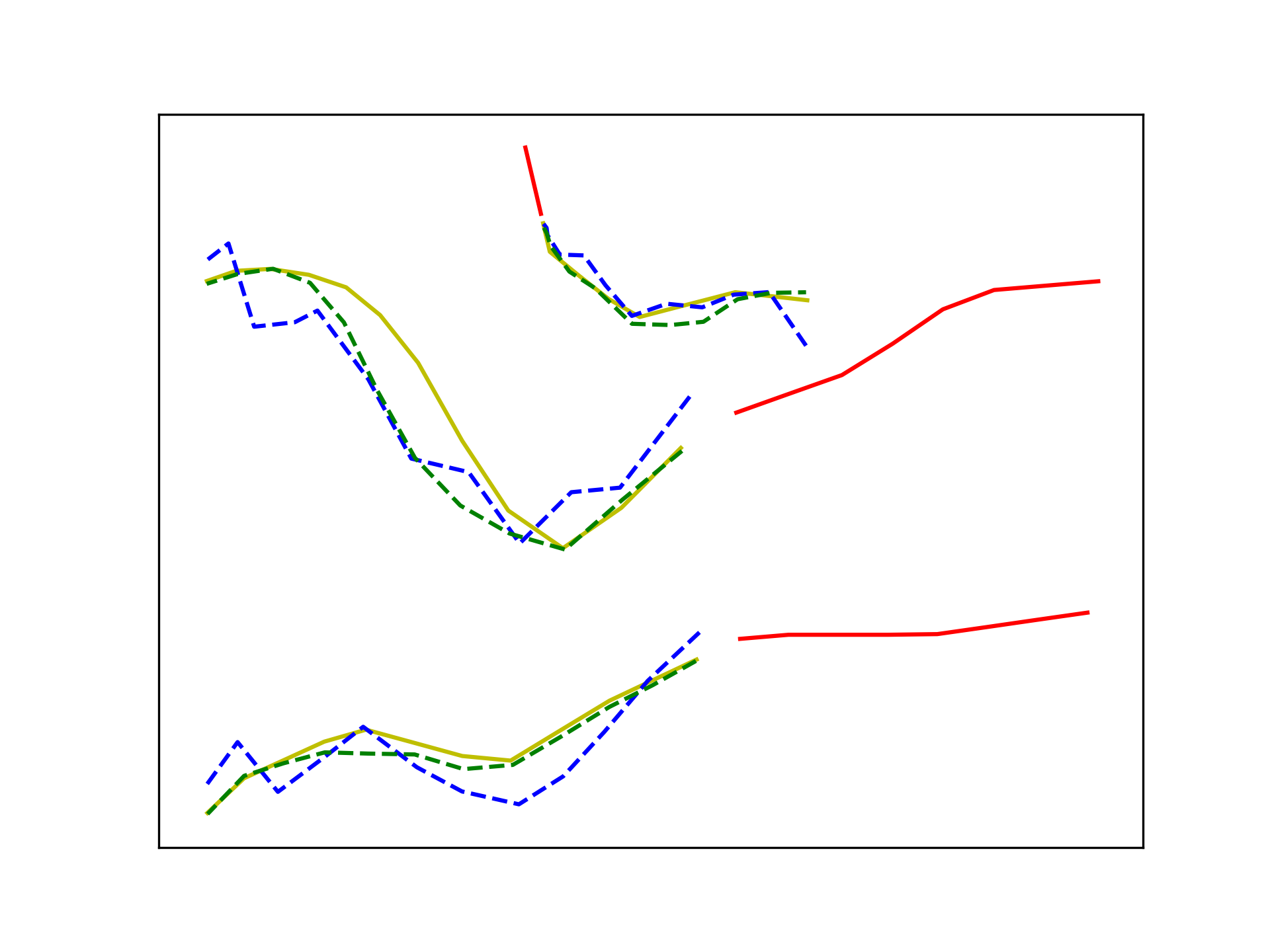}
		\end{minipage}
		\begin{minipage}[b]{0.28\textwidth}
			\includegraphics[width=\textwidth, trim=30 30 30 30, clip]{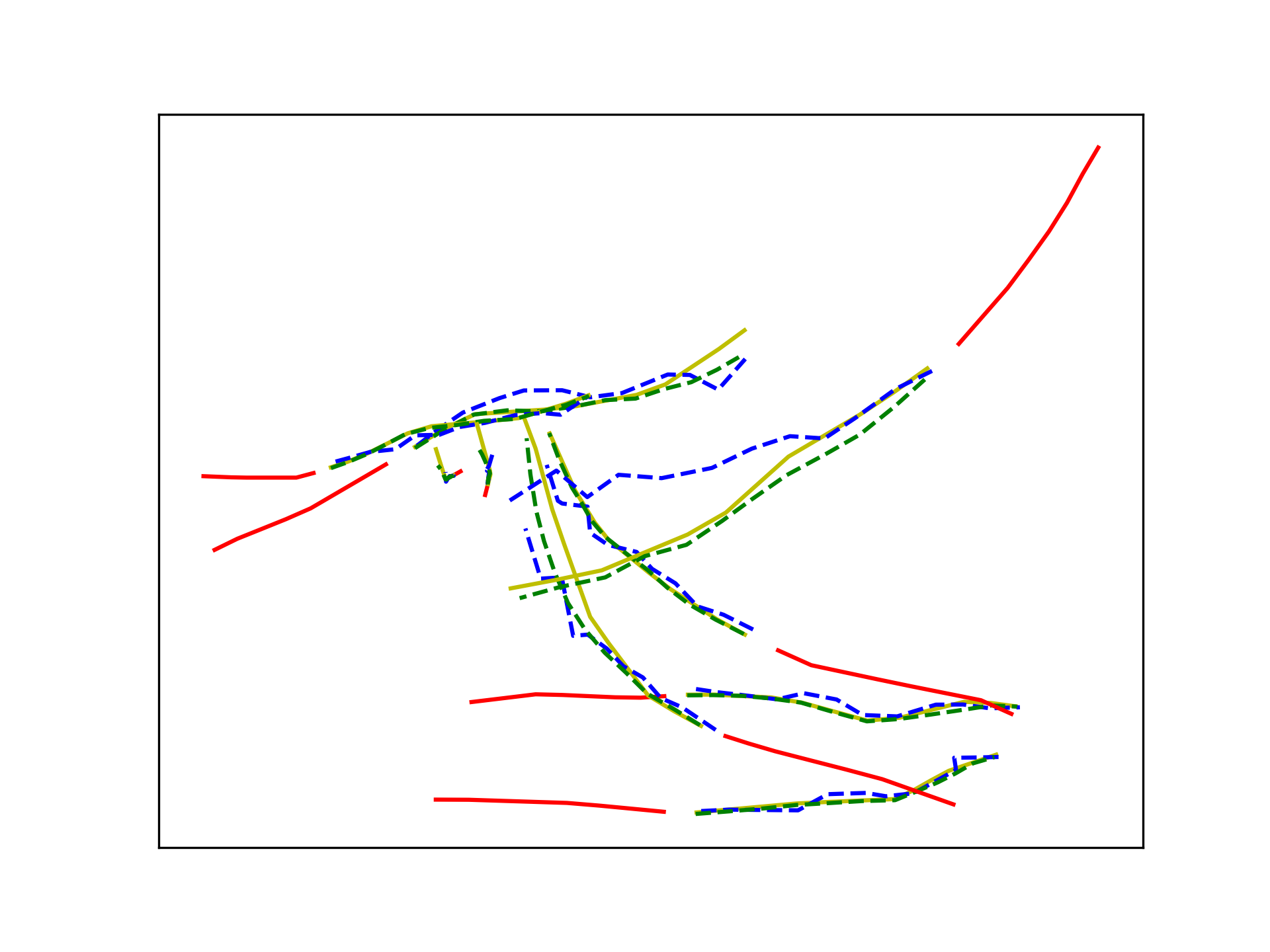}
		\end{minipage}
		\begin{minipage}[b]{0.28\textwidth}
			\includegraphics[width=\textwidth, trim=30 30 30 30, clip]{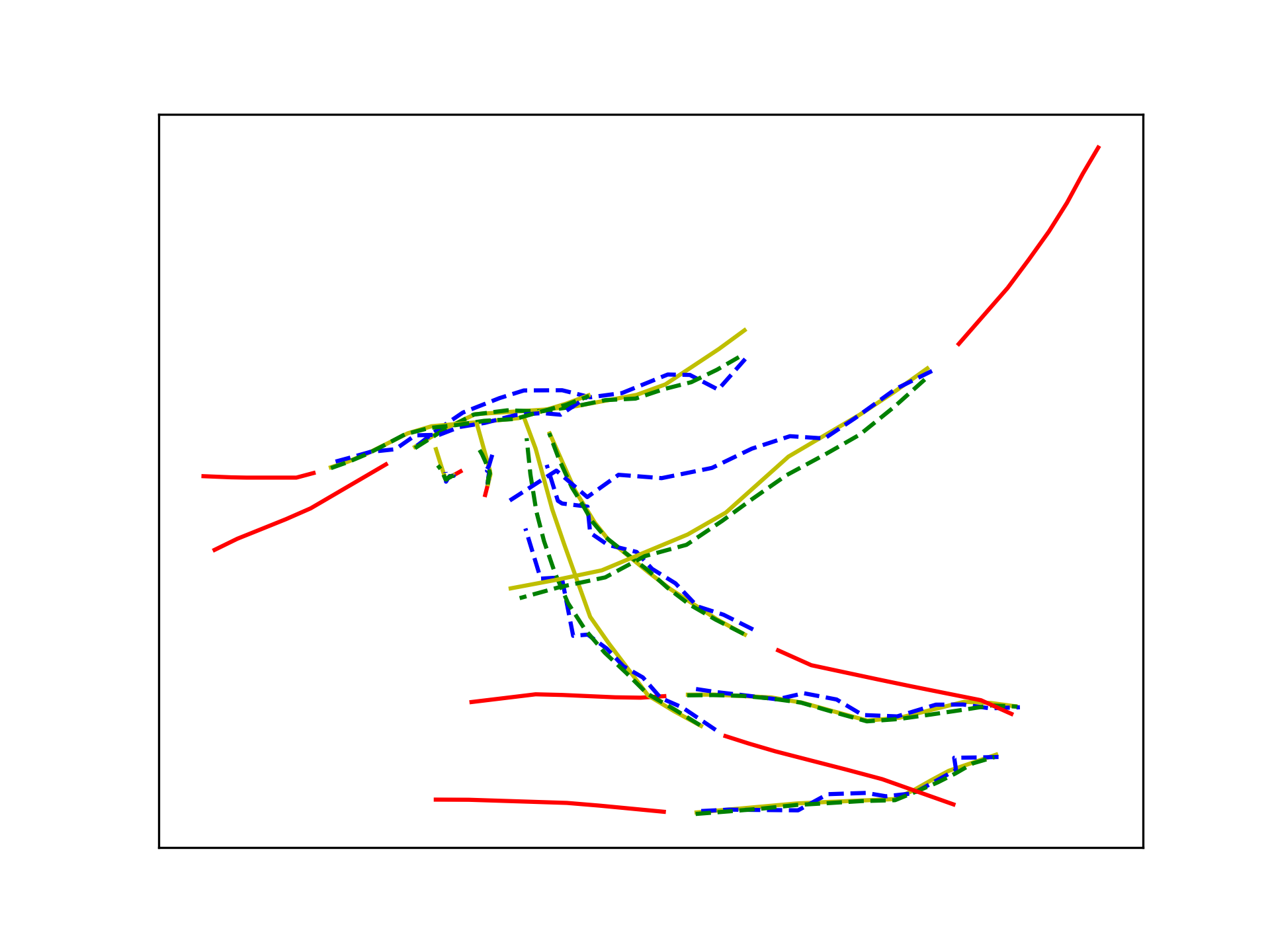}
		\end{minipage}
	}
	\centering
	\caption{Qualitative results of CSR and S-CSR in ETH/UCY dataset. The visualized trajectories are best predictions sampled from 20 trials. The red lines, yellow lines, green dashed lines, and blue dashed lines respectively represents the past trajectories, ground truth future trajectories, predictions of CSR, and predictions of S-CSR.}
	\label{Fig.8}
\end{figure*}

\subsection{Ablation Study}

This experiment is designed to validate the effectiveness of the proposed cascaded CVAE module, socially-aware regression module and slide CVAE module. To validate those modules, a baseline method that consists of $\delta$ CVAEs is introduced for comparison. In the baseline method, $\delta$ future points are respectively predicted by $\delta$ CVAEs that using the original past trajectory as the input, as illustrated in Figure \ref{Fig.5a}.

\subsubsection{Cascaded CVAE Module} The proposed cascaded CVAE module predict future trajectory in a cascaded manner, in which each CVAE sequentially predict a future point using a updated past trajectory that consists of the original past trajectory and the predicted future points before the current time step. To prove the effect of the cascaded CVAE module, it is compared with a baseline method which $\delta$ CVAEs that using the original past trajectory as the input are respectively used to predict $\delta$ future points. As reported in Tables \ref{tab.4} and \ref{tab.5}, the cascaded CVAE module outperforms the baseline by 37.0\%/54.7\% on the ETH/UCY and 47.1\%/49.9\% on the SDD, in terms of the ADE/FDE metrics. This demonstrates that the cascaded CVAE module can extract useful information from the predicted future points, and thus improves the predicition performance of the future trajectory.

\subsubsection{Socially-aware Regression Module} The future trajectory predicted by the cascaded CVAE module may be impractical because the CVAE cannot model interactions among pedestrians. Thus, the socially-aware regression module was introduced to refine the final predictions via social interaction. To prove its effectiveness, the model are respectively trained with and without it. As presented in Tables \ref{tab.4} and \ref{tab.5}, the socially-aware regression module can further boost the performance of the cascaded CVAE module by 17.6\%/4.2\% on the ETH/UCY and 20.4\%/19.7\% on the SDD, in terms of the ADE/FDE metrics, which indicates that this module can extract interaction features from the predicted trajectories to refine the final predictions. Besides, the predictions of the model trained with or without the socially-aware regression module are visualized in Figure \ref{Fig.6}. Figures \ref{Fig.6a}, \ref{Fig.6b} and \ref{Fig.6c} reveal that, compared to the less coordinated trajectories predicted by the cascade CVAE module, more socially compliant trajectories are refined by the socially-aware regression module. Figure \ref{Fig.6d} demonstrates that the the collision problem in trajectory prediction can be improved by the socially-aware regression module using interaxtions between pedestrians.

\subsubsection{Slide CVAE Module} 
The cascaded CVAE module is effectiveness in using the predicted future points to assist in trajectory prediction. But its efficiency is directly affected by two factors, namely the multiple unshared CVAEs and the incremental inputs. To increase the inference speed and decrease the model parameters, a slide CVAE module is proposed to further improve the cascaded CVAE module. As shown in the 6-th and 7-th columns of Tables \ref{tab.4} and \ref{tab.5}, the slide CVAE module improves the cascaded CVAE module by 28.6\%/30.4\% on the ETH/UCY and 43.1\%/45.4\% on the SDD, in terms of the ADE/FDE metrics. Moreover, as depicted in Table \ref{tab.6}, the slide CVAE module decreases the model parameters by 85.9\% and increases the inference speed by 6.3\%.  It is evident that the slide CVAE module has a far better performance and a higher efficiency than the cascaded CVAE module. This is expected, because one trajectory can train one slide CVAE for $\delta$ times. While for the cascaded CVAE modules, one trajectory need to train $\delta$ CVAEs, which has more parameters and is more difficult to fitting.

\subsection{Qualitative Analysis}
Qualitative results of CSR and PECNet on the SDD are visualized in Figure \ref{Fig.7}.  Figures \ref{Fig.7a} and \ref{Fig.7b} present predictions of linear trajectories. The two methods can both produce trajectories that very close to the ground truth. It is expected because the prediction of linear trajectories is easy. Figure \ref{Fig.7c} and \ref{Fig.7d} show predictions of crooked trajectories. In this case, the predictions of CSR are more accurate than that of the PECNet. These results demonstrate that the proposed CSR has better performance in predicting nonlinear trajectories.

Figures \ref{Fig.7e} and \ref{Fig.7f} respectively visualize two easy interaction samples, \textit{i.e.} turning together and straight going together. Both PECNet and CSR can generate trajectories that are close to the ground truth because the interactions between pedestrians are simple. Figure \ref{Fig.7g} and \ref{Fig.7h} visualize predictions of two complex interacions such as crossing and collision avoidance. The predictions of CSR are socially compliant and more reasonable than PECNet. These results demonstrate that the proposed CSR can effectively model the complex interactions between pedestrians.

Qualitative results of CSR and S-CSR on the ETH/UCY dataset are also visualized in Figure \ref{Fig.8}. For each scene, three samples are randomly sampled and visualized. Figure \ref{Fig.8} shows that CSR can capture interactions, and generate socially compliant trajectories in most cases. And compared with the trajectories produced by CSR, the trajectories that more closer to the ground truth future trajectories are predicted by S-CSR.

\section{Conclusion}
In this work, a novel cascaded CVAE with socially-aware rethinking (CSR) method is proposed for pedestrian trajectory prediction. It consists of a cascaded CVAE module and a socially-aware regression module. The cascaded CVAE module predicts the future trajectories in a sequential manner using the updated past trajectory. And the socially-aware regression module extracts interactions features from both the past trajectories and the predicted trajectories to refine the final predictions. Besides, a slide CVAE module is proposed to further improve the model efficiency and performance of the cascade CVAE module using one shared CVAE, in a slidable manner. Extensive experiments on the SDD and ETH/UCY datasets demonstrate the effectiveness of the proposed methods. Further more, the results indicate that the proposed CSR and S-CSR outperforms other state-of-the-art methods by a large margin and achieves a new state-of-the-art performance.

\bibliographystyle{IEEEtran}
\bibliography{aaai22.bib}

\vfill

\end{document}